\documentclass{article}
\PassOptionsToPackage{sort,numbers,compress}{natbib}
\usepackage[preprint]{arxiv}

\usepackage[utf8]{inputenc} % allow utf-8 input
\usepackage[T1]{fontenc}    % use 8-bit T1 fonts
\usepackage{hyperref}       % hyperlinks
\usepackage{url}            % simple URL typesetting
\usepackage{booktabs}       % professional-quality tables
\usepackage{amsfonts}       % blackboard math symbols
\usepackage{nicefrac}       % compact symbols for 1/2, etc.
\usepackage{microtype}      % microtypography
\usepackage[dvipsnames]{xcolor}         % colors
\usepackage{dsfont}
\usepackage{amsmath}
\usepackage{makecell}
\usepackage{graphicx}
\usepackage{caption}
\usepackage{subcaption}
\usepackage{bm}
\usepackage{pifont}% http://ctan.org/pkg/pifont
\newcommand{\cmark}{{\color{OliveGreen}\ding{51}}}%
\newcommand{\xmark}{{\color{red}\ding{55}}}%
\usepackage[symbol]{footmisc}

\usepackage{xspace}
\newcommand*{\eg}{e.g.\@\xspace}

\title{TAPVid-3D: \\ A Benchmark for Tracking Any Point in 3D}

\author{
  \textbf{Skanda Koppula\textsuperscript{1,2}\footnotemark[1]\ },
  \textbf{Ignacio Rocco\textsuperscript{1}\footnotemark[1]\ },
  \textbf{Yi Yang\textsuperscript{1}},
  \textbf{Joe Heyward\textsuperscript{1}},\\
  \textbf{Jo\~ao Carreira\textsuperscript{1}},
  \textbf{Andrew Zisserman\textsuperscript{1,3}},
  \textbf{Gabriel Brostow\textsuperscript{2}}, 
  \textbf{Carl Doersch\textsuperscript{1}}
  \\
  \textsuperscript{1}Google DeepMind \quad
  \textsuperscript{2}University College London \quad
  \textsuperscript{3}University of Oxford \quad
}

\begin{document}

\maketitle

\footnotetext[1]{Equal contribution. Corresponding author: \href{mailto:skandak@google.com}{\texttt{skandak@google.com}}}

\begin{abstract}
We introduce a new benchmark, {\em TAPVid-3D}, for evaluating the task of long-range Tracking Any Point in 3D (TAP-3D). While point tracking in two dimensions (TAP) has many benchmarks measuring performance on real-world videos, such as TAPVid-DAVIS, three-dimensional point tracking has none. To this end, leveraging existing footage, we build a new benchmark for 3D point tracking featuring 4,000+ real-world videos, composed of three different data sources spanning a variety of object types, motion patterns, and indoor and outdoor environments. To measure performance on the TAP-3D task, we formulate a collection of metrics that extend the Jaccard-based metric used in TAP to handle the complexities of ambiguous depth scales across models, occlusions, and multi-track spatio-temporal smoothness. We manually verify a large sample of trajectories to ensure correct video annotations, and assess the current state of the TAP-3D task by constructing competitive baselines using existing tracking models. We anticipate this benchmark will serve as a guidepost to improve our ability to understand precise 3D motion and surface deformation from monocular video. Code for dataset download, generation, and model evaluation is available at \url{https://tapvid3d.github.io/}.
\end{abstract}

\section{Introduction}

\begin{figure}[h]
  \centering
  \includegraphics[width=\linewidth]{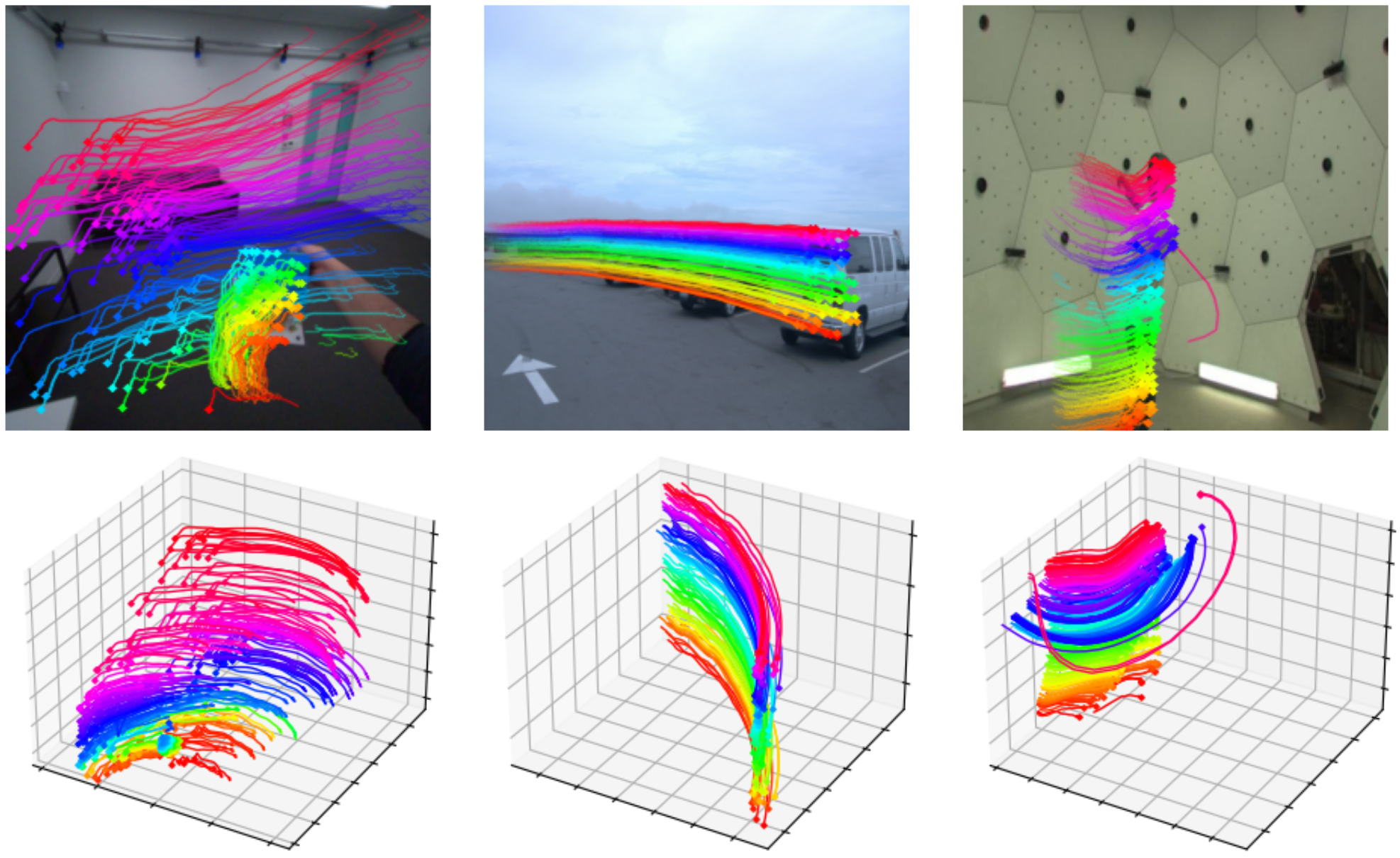}
  \caption{Random samples from TAPVid-3D: on the top row, we visualize the point trajectories projected into the 2D video frame; on the bottom row, we visualize the metric 3D point trajectories. From left to right, we show one example from each constituent data source: ADT, DriveTrack and Panoptic Studio.}
  \label{fig:teaser}
\end{figure}

For robots, humans, and other agents to effectively interact with the physical 3D world, it is necessary to understand a scene's structure and dynamics. This is a key ingredient to any general embodied intelligence: the ability to learn a world model to understand and predict the structure and motion of arbitrary scenes. It is attractive to leverage the vast amounts of monocular video data that is available cheaply, and use such signals to understand the geometry and 3D motion in real-world footage. But how well can current perception algorithms actually do this?

The field has seen many efforts to measure 3D and motion understanding from videos, each contributing a part of the overall goal. For example, monocular depth estimation is a widely recognized task~\cite{casser2019depth,luo2020consistent,yang2024depth}. However, success in depth estimation alone does not reveal whether the model understands how surfaces move from one frame to the next, and may not respect temporal continuity. For instance, a basketball spinning on its axis is often not visible in a sequence of depth maps. On the other end of the generality spectrum, 3D pose tracking, \eg for rigid objects~\cite{hai2023rigidity, lourakis2013model} and people~\cite{sminchisescu20083d, goel2023humans}, evaluates precise motion, but requires a known 3D model of each object with articulations.  Building parametric models and pose estimation methods for all animal classes is infeasible, much less for all the objects that a robot might encounter in, for example, an unseen, busy construction site.

An alternative and more general approach for dynamic scene understanding observes that the world consists of particles, each of which individually follows a 3D trajectory through space and time~\cite{wang2023tracking,xiao2024spatialtracker}. Measuring the motion of these points provides a way to measure 3D motion \textit{without requiring any 3D shapes to be known a priori}. To this end, in TAPVid-3D, we focus on providing the community with a benchmark consisting of real world videos and three-dimensional point tracking annotations, spanning a wide variety of objects, scenes, and motion patterns.

Prior work on 2D understanding has shown that this kind of point-wise motion can be extremely valuable: both optical flow and long term two-dimensional point tracking (TAP) tasks have been applied to video editing~\cite{yu2023videodoodles}, controllable video generation~\cite{wang2023motionctrl}, robotic manipulation~\cite{vecerik2023robotap, abbeel}, and more~\cite{chao2014survey, huang2020improved}. TAP is an  occlusion-aware, temporal extension of optical flow, which itself has a 3D extension called \textit{scene flow}~\cite{mayer2016large, menze2015object,ma2019deep}. While useful, scene flow suffers from the same challenges as optical flow, namely, that it captures instantaneous motion and does not evaluate correct, occlusion-aware association of pixels over long sequences. A new three-dimensional TAP benchmark would provide a way to measure progress on many of these tasks: our TAPVid-3D benchmark targets evaluating general motion understanding, for models performing both dense and sparse particle tracking, in two and three dimensions.

Unfortunately, all currently used evaluations for TAP on real-world videos assess only 2D tracking ability (\eg the TAP-Vid suite~\cite{doersch2022tap}, BADJA~\cite{biggs2019creatures},  CroHD~\cite{sundararaman2021tracking}, and JHMDB~\cite{Jhuang:ICCV:2013}), and cannot evaluate the performance of 3D point tracking due to lack of access to the ground-truth metric position trajectories. While evaluations based on synthetic environments, like Kubric~\cite{greff2022kubric}, RGB-Stacking~\cite{doersch2022tap}, and Point Odyssey~\cite{zheng2023pointodyssey}, could potentially provide 3D point tracking annotations, these introduce a significant domain gap with real-world scenes and are therefore less representative of model performance on real-world tasks.

Many applications stand to benefit from direct evaluation of three-dimensional point tracking capabilities and improvements to such models. For example, robotic manipulation tasks are likely to be easier with accurate 3D motion estimates, to understand the changing relative world position of the gripper, any objects, and the background. Video generation models would be more useful if creators were able to condition on exact motion tracks describing the 3D movement of both objects and the camera, as a director would do on a stage. Standard scene understanding tasks like velocity estimation, motion prediction, and object parts segmentation are simpler given the 3D motion tracks of individual points. Many visual odometry, mapping, and structure from motion pipelines rely on accurate 3D correspondences; with the ability to track 3D point correspondences from \textit{any} pixel, such pipelines could be made more robust and accurate, even with many moving objects. Overall, the task of three dimensional point tracking provides a strict superset of information as compared to its 2D counterpart, is likely to be more useful in downstream applications, and provides a greater test of physical world motion understanding.

To this end, we introduce TAPVid-3D: a \emph{real-world} benchmark for evaluating the Tracking Any Point in 3D (TAP-3D) task. The benchmark contributes: (1) a unification of three distinct real-world video data sources, with pipelines to generate, standardize, and validate consistent ground-truth $(x,y,z)$ 3D trajectories and occlusion information, (2) formalization of the TAP-3D task, with new metrics to measure accuracy of 3D track estimation, and (3) an assessment of the current state of TAP-3D, formed from the first real-world video evaluations of the nascent set of early 3D TAP models. Code for dataset download, generation, and model evaluation is available at \url{https://tapvid3d.github.io/}.

\section{Related work}

\paragraph{Tracking Any Point.} In recent years, long-range tracking of local image points has been formalized as the Tracking Any Point task (TAP) and evaluated using the, now standard, TAPVid benchmark~\cite{doersch2022tap}, among others. Success on TAP consists of tracking the 2D trajectory of any given $(x,y)$ \emph{query point}, defined at a particular frame $t$, throughout the rest of a given video. By definition, the query point is considered visible at the query frame, and associated to the material point of the scene which is observed at $(x,y,t)$.  However, this material point may become occluded or go out of the image boundaries. To handle this, TAP models also must predict a binary visibility flag $v$ for each timestamp of the video. The tracked $(x,y)$ positions and visibility estimates are scored jointly using an all-encompassing average Jaccard metric.
Most current TAP models~\cite{harley2022pips,doersch2022tap,doersch2023tapir,karaev2023cotracker,doersch2024bootstap} are limited to tracking in 2D pixel space. Recently, some works have started exploring the extension of the TAP problem to 3D (TAP-3D)~\cite{xiao2024spatialtracker,wang2024scenetracker}. However, most TAP benchmarks containing real-world videos, such as TAPVid-DAVIS~\cite{doersch2022tap}, Perception Test~\cite{patraucean2024perception}, CroHD~\cite{sundararaman2021tracking} and BADJA~\cite{biggs2019creatures}, don't have 3D annotations, and therefore evaluations are still performed on the 2D tracking task. Concurrent to our work, Wang et al.~\cite{wang2024scenetracker} introduced both a synthetic (LSFOdyssey) and a real benchmark (LSFDriving) for evaluating TAP-3D. However, their real benchmark is limited to the driving domain and only contains 180 test clips with 40 frames each. Our proposed TAPVid-3D benchmark is substantially larger and more diverse, containing 4000+ clips with durations between 25 and 300 frames.

\paragraph{Scene flow estimation.} The scene flow estimation problem, introduced by Vedula et al.~\cite{vedula1999three}, seeks to obtain a dense, instantaneous, 3D motion field of a 3D scene, analogously to the way optical flow estimates a dense 2D motion field across consecutive frame pairs of a 2D video. The TAP-3D task is related to the scene flow problem in a similar way in which the TAP task is related to the optical flow problem. While scene flow seeks to obtain dense instantaneous motion estimation, TAP-3D seeks to obtain longer-range tracking, spanning tens or hundreds of frames. Furthermore, TAP-3D does not seek to produce dense fields of tracks, but is rather interested in tracking a sparse set of query points, which is more computationally tractable. From work in TAP, we have observed that having motion representations with longer temporal context is useful for downstream tasks such as robotic manipulation, while having sparser spatial coverage is sufficient for many tracking applications.

\paragraph{Pose estimation.} Closely related to point tracking is pose estimation and keypoint tracking. Many benchmarks have been proposed for 2D and 3D pose estimation~\cite{zhang2013actemes, haque2016towards, li2019crowdpose, von2018recovering}. 3D pose estimation tasks and benchmarks largely focus on specific categories of moving objects, and for objects that are articulated: e.g. humans~\cite{von2018recovering, johnson2010clustered}, hands~\cite{tompson14tog}, animals~\cite{yu2021ap, mathis2021pretraining}, and even jointed furniture~\cite{lim2013parsing}. For general motion and scene understanding, we aim to learn motion estimation across any object or scene pixel, expanding the generality of the task.

\paragraph{Static scene reconstruction.} 
Static scene reconstruction, a fundamental problem in computer vision, has been advanced through techniques like Structure-from-Motion (SfM) and monocular depth estimation. Significant contributions include COLMAP~\cite{schonberger2016structure} for state-of-the-art reconstructions and MVSNet~\cite{yao2018mvsnet}, which enhances multi-view stereo depth estimation with deep learning. These studies collectively advance robust and precise static scene reconstruction. Evaluation of the local features and depth are crucial for static scene reconstruction methods. ~\cite{schonberger2017comparative} provided a comparative evaluation of hand-crafted and learned features, while MegaDepth~\cite{li2018megadepth} improved single-view depth prediction using large scale multi-view Internet photo collections. Despite significant progress, static scene reconstruction struggles with dynamic environments, highlighting the need for dynamic scene methods. 

\paragraph{Dynamic scene reconstruction.}
3D reconstruction of dynamic scenes and objects is a widely studied problem in computer vision. Over the years, several methods have proposed solutions to this problem, starting with Non-rigid Structure-from-Motion methods (NRSfM)~\cite{bregler2000recovering,nrsfmtraj}. While these methods have shown some success modelling simple motions like facial expressions and skeletal motions, they fail to generalize to arbitrary motions. More recently, deep-learning based methods have been used to perform 3D reconstruction of dynamic scenes. One line of works exploit Monocular Depth Estimation models~\cite{ranftl2020towards} and performs test-time optimization on each given video to obtain smoother reconstructions, under the assumption that the frame rate of the camera is high with respect to the speed of the depicted object (quasi-static scene assumption)~\cite{luo2020consistent, zhang2021consistent, kopf2021robust}. While these methods are more general than the classic NRSfM counterparts, they still require costly test-time optimization and fail to model rapid motions. Other lines of work attempt to fit a neural-scene representation to each video, such as neural-radiance fields~\cite{li2021neural, li2023dynibar} or 3D Gaussian Splatting~\cite{yang2023real}. However, these methods require a costly per-video optimization, and typically make smoothness and local-rigidity assumptions about the motion of the points in the scene. We believe the development of TAP-3D methods should significantly help for the problem of dynamic scene reconstruction, as these models can run in a feed-forward manner, without requiring test-time optimization, and don't need any explicit motion prior assumptions as they can learn these from data.

Table~\ref{table:unique} summarizes the focus areas of measurement for common scene understanding benchmarks. On the bottom row is the proposed TAP-3D task, and corresponding benchmark, TAPVid-3D, which brings to the table a more complete test of dynamic scene understanding in one simple evaluation.

\begin{table}[tb]
\resizebox{\textwidth}{!}{
\renewcommand{\arraystretch}{1.1}
\begin{tabular}{@{}rcccccc@{}}
\toprule
\begin{tabular}[c]{@{}r@{}}\\
\textbf{Benchmark Type}
\end{tabular}  & \makecell{\textbf{Long Term} \\ \textbf{Continuity}} & \textbf{3D} & \makecell{\textbf{Pixel Level} \\ \textbf{Occlusion}} & \makecell{\textbf{Pixel Level} \\ \textbf{Motion}} & \makecell{\textbf{Non-rigid} \\ \textbf{Surfaces}} & \makecell{\textbf{No Prior} \\ \textbf{3D Model}} \\ \midrule

Monodepth~\cite{song2015sun, butler2012naturalistic, li2018megadepth}            & \xmark & \cmark & \xmark & \xmark & \cmark & \cmark \\
2D Point Tracking (TAP)~\cite{doersch2022tap}               & \cmark & \xmark & \cmark & \cmark & \cmark & \cmark \\
Scene flow~\cite{khatri2024can, wilson2023argoverse}           & \xmark & \cmark & \xmark & \cmark & \cmark & \cmark \\
3D Pose Tracking~\cite{liu2022hoi4d, von2018recovering} & \cmark & \cmark & \xmark & \cmark & \xmark  & \xmark \\
3D Object Box Tracking~\cite{sun2020waymoopen, liao2022kitti}         & \cmark & \cmark & \xmark & \xmark & \xmark  & \cmark \\
\textbf{3D Point Tracking (TAP-3D)}              & \cmark & \cmark & \cmark & \cmark & \cmark & \cmark \\
\bottomrule
\end{tabular}
}
\vspace{2mm}
\caption{The proposed TAPVid-3D benchmark provides a unique set of characteristics, not covered in previous tasks or benchmarks. It extends the temporal continuity, occlusion modeling, and motion estimation capabilities of TAP benchmarks into 3D. Each row describes aspects tested in each task.} \vspace{-5mm}
\label{table:unique}
\end{table}

\section{TAPVid-3D\label{sec:tapvid3d}}

We build a real-world benchmark for evaluating Tracking Any Point in 3D (TAP-3D) models. To do this, we leveraged three publicly available datasets: (i) Aria Digital Twin~\cite{pan2023aria}, (ii) DriveTrack~\cite{balasingam2023drivetrack, sun2020waymoopen} and (iii) Panoptic Studio~\cite{hanbyuljoo2019panoptic}. These data sources span different application domains, environments, and video characteristics, deriving ground truth tracking trajectories from different sensor types. For instance, Aria Digital Twin is a dataset of egocentric video, and is more close to bimanual robotic manipulation problems, where the camera is robot mounted and sees the actions from a first person view. DriveTrack features footage captured from a Waymo car navigating outdoor scenes, akin to tasks in robotic navigation and outdoor, rigid-body scene understanding tasks. Finally, Panoptic Studio captures third-person views of people performing diverse actions within an instrumented dome. It presents complex human movement which aligns more closely with NRSfM-adjacent tasks. We believe that, combined, these data sources present a diverse and comprehensive benchmark of TAP-3D capabilities for many potential downstream tasks. We describe our pipeline to extract ground truth metric 3D point trajectories from each source in the next sections, with samples in Figure~\ref{fig:teaser}.

Table~\ref{table:statistics} shows the summary statistics across the entire dataset, and for the dataset subdivisions corresponding to each constituent data source. As there are comparatively a large number of videos (for reference, the commonly used TAPVid-DAVIS has 30 videos, whereas TAPVid-3D has two orders of magnitude more), we release two splits: a \texttt{minival} split, with 50 videos from each data source, and a \texttt{full\_test} split, containing all videos in the benchmark, without the \texttt{minival} videos. The \texttt{minival} is intentionally lightweight, and likely most useful for online evaluation during training.

\subsection{Aria Digital Twin} This split employs real videos captured with the Aria glasses~\cite{ariaglasses} inside different recording studios, which mimic household environments. Accurate digital replicas of these studios, created in 3D modeling software, are used to obtain pseudo-ground truth annotations of the original footage. This includes annotations such as segmentation masks, 3D object bounding boxes and depth maps. We leverage these annotations to extract 3D trajectories from given 3D query points. In particular, given a video $V=\{I_t\}_{t=1,\dots, T}$ with $T$ frames and $W\times H$ spatial resolution, with corresponding object segmentation masks $\{S_t\}\subset \mathbb{Z}^{W\times H}$, and depth maps $\{D_t\} \subset \mathbb{R}^{W\times H}$, extrinsic world-to-camera poses\footnote{We use the notation $P^a_b$ to represent the SE(3) transform from coordinate frame $a$ to frame $b$.} $\{(P^{w}_{cam})_t\}\subset \mathbb{R}^{3\times 4}$, camera intrinsics $K\in \mathbb{R}^{3\times 3}$, and a query point $q = (x_q,y_q,t_q)$, we first extract the query point's 3D position $Q_{cam}$\footnote{We use the notation $Q_{cam}$ and $Q_{obj}$ to denote the position of the 3D point $Q$ in the camera and object coordinate frames, respectively.} in the camera coordinate frame, with
 
\begin{equation}
    (Q_{cam})_{t_q} = K^{-1}  (x_q, y_q, 1)^T \cdot D_{t_q}(x_q,y_q).
    \label{eq:unproject}
\end{equation}

Additionally, we obtain the object ID of the query point from the segmentation mask $q_{id} = S_{t_q}(x_q, y_q)$, and use it to retrieve the 3D object pose of the query object $(P^{w}_{obj})_{t_q}$, which converts points from world coordinate frame to the object coordinate frame. This allows us to compute the query point position in the object's frame of reference as
 
\begin{equation}
    Q_{obj} = (P^{w}_{obj})_{t_q}(P^{cam}_{w})_{t_q}(Q_{cam})_{t_q}.
    \label{eq:q_obj}
\end{equation}

In this way, we \emph{fix} the query point to the corresponding object, and then obtain its track across the whole video by leveraging the object's pose annotation. For any timestamp $t$, the position of the query point can be then obtained by
 
\begin{equation}
(Q_{cam})_{t} = (P^{w}_{cam})_{t} (P^{obj}_{w})_{t} Q_{obj}.
    \label{eq:q_cam_t}
\end{equation}

To compute the visibility $v$ of the query point at any time, we first employ a pretrained semantic segmentation model to obtain the semantic masks of the operator's hands $\{H_t\}$, as these are not modelled in the digital replica. Then, we compute the visibility $v$ by verifying that the query point's depth is close to the observed depth, and it does not lie on the hands segmentation mask $H$, so

\begin{equation}
    v_t= \mathds{1}(\vert (Z(Q_{cam}) - D_t(u,v)\vert < \delta) \cdot (1-H_t(u,v)),
    \label{eq:visibility}
\end{equation}

where $(u,v) = \Pi_K((Q_{cam})_t)$ is the projection of the query point $(Q_{cam})_t$ to the image plane according to the given camera intrinsics $K$, and $Z((x,y,z)) = z$ is the function that extracts the z-component of a 3D point. This approach allows us to compute the 3D trajectory $\{(Q_{cam})_t\}$ and visibility flag $\{v_t\}$ of the query point across the whole video. 

\subsection{DriveTrack}

The DriveTrack split is based on videos from the Waymo Open dataset~\cite{sun2020waymoopen}, and the 2D point trajectory pipeline used in DriveTrack~\cite{balasingam2023drivetrack}. In particular, each frame $I_t$ in a video sequence $V$ has a corresponding, time-synchronized point cloud $\{C_t\}$ from the Waymo car's LIDAR. The subset of points that correspond to a randomly selected, single, tracked vehicle $(Q_{cam})_{t_s}$ are subselected from the entire point cloud $(Q_{cam})_{t_s} \subset {C}_{t_s}$ at a certain sampling time $t_s$ using a manually-annotated 3D bounding box around the chosen object. These object-specific points are then tracked across the whole video using: (i) a vehicle rigidity assumption, and, (ii) the pose and position of the object's annotated 3D bounding box through the entire video sequence. Specifically, the tracked points in object coordinate frame $Q_{obj}$ are first computed using~\eqref{eq:q_obj}, and then the trajectories in camera coordinate frame $\{(Q_{cam})_t\}$ are obtained using~\eqref{eq:q_cam_t}.

Visibility flag is estimated by first computing the dense depth map $D_t$ of the corresponding camera video frames through interpolation of sparse LIDAR values as in \cite{balasingam2023drivetrack}. This is compared to the point's depth computed from the 3D point trajectory given by $(Q_{cam})_t$, as in the first term of~\eqref{eq:visibility}. If the point depth (distance from the camera center to the query point) is greater than the depth provided by the depth map (with a 5\% relative threshold margin), it is marked as not visible. 

Finally, to determine the 2D query points $q=(x_q, y_q, t_q)$ we sample $t_q$ uniformly among the visible timestamps $v_t$, and then obtain $(x_q,y_q) = \Pi_K((Q_{cam})_{t_q})$. 

\subsection{Panoptic Studio}
The original Panoptic Studio dataset~\cite{hanbyuljoo2019panoptic} consists in video sequences captured inside a recording dome using stationary cameras, and depicting different actors performing various actions such as passing a ball or swinging a tennis racket, featuring complex non-rigid motions. To obtain 3D trajectories, we leverage the pretrained dynamic 3D reconstructions from Luiten et al.~\cite{luiten2023dynamic}. These have been obtained by first performing a rigid 3D reconstruction through 3D Gaussian Splatting~\cite{kerbl20233dgs}, fitting a set of 3D Gaussians $\{(\mu_i,\Sigma_i)_{t_0}\}_{i=1,\dots,N}$ to the first timestamp $t_0$ of each sequence using the multiple cameras available in the dome. Then, these Gaussians are displaced and rotated in a as-rigid-as-possible manner to model the motion occurring in the subsequent frames of the video. For more details, please refer to~\cite{luiten2023dynamic}. Using these pretrained splatting models, we render pseudo-ground-truth depth maps $\{D_t\}$ for each sequence. Then, given an query point $q=(x_q,y_q,t_q)$, we unproject the point to 3D following~\eqref{eq:unproject}, obtaining $(Q_{cam})_{t_q}$, and retrieve the index $i^*$ of the closest Gaussian center at that time using the poses $\{(P^{w}_{cam})_t\}$, so that $i^* = \underset{i=1,\dots,N}{\text{argmin}} \|(Q_{cam})_{t_q} - (P^{w}_{cam})_{t_q}(\mu_i)_{t_q}\|_2$.

Note that due to the distance between $(Q_{cam})_{t_q}$ and $(P^{w}_{cam})_{t_q}(\mu_{i^*})_{t_q}$, their projections onto the image plane will not match exactly. To account for this difference, we adjust the query point's 2D position as
    $(x_q, y_q) = \Pi_K (P^{w}_{cam})_{t_q}(\mu_{i^*})_{t_q}$.
Then, the query point's 3D track come by following the motion of the $i^*$ Gaussian center, so $
 (Q_{cam})_t  \equiv (P^{w}_{cam})_{t}(\mu_{i^*})_t$.

Visibility $\{v_t\}$ is estimated as in~\eqref{eq:visibility} (omitting the second term), by comparing the depth of our 3D query point with the observed depth from $D_t$. We only track points across the foreground deforming characters, as tracking background points is trivial given that the cameras in this dataset are stationary.

\subsection{Data Cleanup and Validation}

While the aforementioned procedures generally produce high quality trajectories, small amounts of noise from the underlying dataset sources can cause issues in a small fraction of sequences. These minor inaccuracies, for example, can be caused by small misalignement between Aria Digital Twin synthetic annotations and the real world video, LIDAR sensor noise, insufficiently constrained Gaussian splats, or numerical error. We minimized these errors through automated methods, and then manually checked a sampling of the videos to ensure accuracy.

Firstly, since trajectories are descriptors of surface motion, their motion should be localized to their associated object. We use instance segmentation models to generate object masks for each frame~\cite{kirillov2023segment}, filtering out errant trajectories that exceed these boundaries when not occluded. In DriveTrack specifically, the tracked bounding box is an approximation to the true object mesh, but such errors are fixed with tight segmentation masks (trimming 2-3\% of initial trajectories).

Secondly, we observed that some trajectories have a `flickering' visibility flag. This is not unique to TAPVid-3D, as we notice this in the widely used Kubric~\cite{greff2022kubric} and DriveTrack~\cite{balasingam2023drivetrack}. However, to mitigate this in our dataset, we oversample trajectories in the annotation generation pipeline, and filter trajectories whose visibility state changes more times than 10\% of the total number of video frames.  More details can be found in the appendix.

\begin{table}[tb]
\footnotesize
\setlength{\tabcolsep}{2pt}
{\centering
\begin{tabular}{@{}rccccccc@{}}
\toprule
\textbf{Dataset split}  & \textbf{\#clips (minival)}  & \textbf{\#trajs per clip} & \textbf{\#frames per clip} & \textbf{\#videos} & \textbf{\#scenes} & \textbf{resolution}        & \textbf{fps}   \\ \midrule
Aria Digital Twin & $1956$ ($50$)  & $1024$      & $300$         & $215$    & $2$    & $512\times 512$   & $30$    \\
DriveTrack                                                   & $2457$ ($50$)  & $256$       & $25-300$      & $2457$   & $252$   & $1920\times 1280$ & $10$    \\
Panoptic Studio                                              & $156$ ($50$)    & $50$        & $150$         & $156$    & $1$  & $640\times 360$   & $30$    \\ \midrule
TAPVid-3D                                                    & $4569$ ($150$)    & $50-1024$   & $25-300$      & $2828$   & $255$    & Multiple          & $10 / 30$ \\
\bottomrule
\end{tabular}
}
\vspace{2mm}
\caption{Overview statistics of the TAPVid-3D benchmark dataset, for all three constituent splits and in total. Clips in the benchmark are temporally sampled from their original video.}
\vspace{-6mm}
\label{table:statistics}

\end{table}

\subsection{Metrics}
To accompany the TAPVid-3D dataset, we adopt and extend the metrics used in TAP~\cite{doersch2022tap} to the 3D tracking domain. These metrics measure the quality of the predicted 3D point trajectories (APD), the ability to predict point visibility (OA), or both simultaneously (AJ).

The \textbf{APD ($< \delta^{x}_{avg}$)} metric measures the average percent of points within $\delta$ error.  If $\hat{P}^i_t$ is the $i$'th point prediction at time $t$, $P^i_t$ is the corresponding ground truth 3D point, and $v^i_t$ is the ground-truth point visibility flag, then:

\begin{equation}
\text{APD}_{3D} \equiv \frac{1}{V}\sum_{i,t} v^i_t \cdot \mathds{1}(\|\hat{P}^i_t-P^i_t\| < \delta_{3D}(P^i_t)),
\label{eq:apd}
\end{equation}

 where $\mathds{1}(\cdot)$ is the indicator function, $\| \cdot \|$ is the Euclidean norm, $V=\sum_{i,t} v^i_t$ is the total number of visibles, and  $\delta_{3D}(P^i_t)$ is the threshold value.

 The value of this threshold is \textit{relative to ground-truth depth}; and is defined by \emph{unprojecting} a pixel threshold $\delta_{2D} \subset \{1 ,2, 4, 8, 16\}$ to 3D space using the camera intrinsic parameters: $\delta_{3D}(P^i_t)=Z(P^i_t)\cdot \delta_{2D} / f$, where $f$ is the camera focal length. We argue that points that are closer to the camera are of higher importance than those that are far away, which is enforced by the definition of $\delta_{3D}$\footnote{A similar depth-adaptive threshold approach is used in Monodepth literature~\cite{ranftl2020towards}.}. Note that by using this definition, the $\text{APD}_{3D}$ metric defined in~\eqref{eq:apd} is numerically equivalent to the $\text{APD}_{2D}$ metric from~\cite{doersch2022tap} when the point estimations $\hat{P}^i_t$ all have correct depths.

In addition, it is important to distinguish between occluded and visible points, because downstream algorithms often want to rely exclusively on predictions which are based on visual evidence, rather than on the guesses that might be very wrong.  To this end, we adopt the occlusion accuracy metric (\textbf{OA}) from~\cite{doersch2022tap}, which computes the fraction of points on each trajectory where $\hat{v}^i_t=v^i_t$, where $\hat{v}^i_t$ is the model's (binary) visibility prediction.  

Finally, we define \textbf{3D-AJ}, 3D Average Jaccard, following TAP, which combines OA and $\text{APD}_{3D}$.  The AJ metric calculates the number of \emph{true positives} (number of points within the $\delta_{3D}$ threshold, predicted correctly to be visible), divided by the sum of \emph{true positives} and \emph{false positives} (predicted visible, but are occluded or farther than the threshold) and \emph{false negatives} (visible points, predicted occluded or predicted to exceed the threshold).  Mathematically, it is defined as:

\begin{equation}
\text{AJ}_{3D} \equiv \frac{\sum_{i,t} v^i_t \, \hat{v}^i_t \, \alpha^i_t}{\sum_{i,t} v^i_t + \sum_{i,t} \bigl((1-v^i_t)\,\hat{v}^i_t\bigr) + \sum_{i,t}\bigl(v^i_t \, \hat{v}^i_t \, (1-\alpha^i_t) )\bigr)},
\end{equation}

where $\alpha^i_t = \mathds{1}(\|\hat{P}^i_t-P^i_t\| < \delta_{3D}(P^i_t))$ indicates whether the point prediction is below the distance threshold.
Note the relationship between this and prior metrics: if the depth estimates for a given point are perfect, then this metric reduces to 2D-AJ, as there will be no depth errors. It is also related to a common Monodepth's metric $\delta < 1.25^k$, which also places a hard threshold on the \textit{relative} depth of the estimated point w.r.t. the ground truth; if the video is a single frame (occlusion-free) and the query points are dense, then there should be no 2D errors, and our metric will behave like a Monodepth metric.  For general sequences, however, the algorithm must output correct tracking \textit{and} correct depth---i.e., a full 3D trajectory---in order to be considered correct by our metric.

Finally, one additional complication is \textit{scale ambiguity}. As in monocular depth literature~\cite{ranftl2020towards}, we globally re-scale predictions to match ground truth, before computing the metrics. We do this by multiplying predictions $\hat{P}^i_t$ by the median of the depth ratios $\|P^i_t\|/\|\hat{P}^i_t\|$ over all points and frames, which we call \textit{global median} rescaling.  Note, however, that some algorithms may be more adept at estimating the relative depth of \emph{individual} points: algorithms trained in simulation, for instance, may be able to estimate the depth change for a single point just by analyzing the frequencies.  
With a slight change to the rescaling, we can accomodate such methods even when they don't produce consistent scale between points.  Therefore, we define the \textit{per-trajectory} rescaling, which rescales each track $P^i$ separately multiplying by $\|P^i_{t_q}\|/\|\hat{P}^i_{t_q}\|$, where $t_q$ is the query timestamp.  Finally, we explore a hybrid approach (which considers scaling in local neighborhoods to better account for object-object interactions) in the supplementary material. 3D-AJ and related metrics are computed using a fixed $256 \times 256$ video resolution, as done with TAPVid-2D \cite{doersch2022tap}.
\vspace{-2mm}

\begin{figure}[t]
\centering
\begin{subfigure}{0.24\textwidth}
  \centering
  \includegraphics[clip, height=6.2cm, trim=0cm 0.3cm 0cm 0cm]{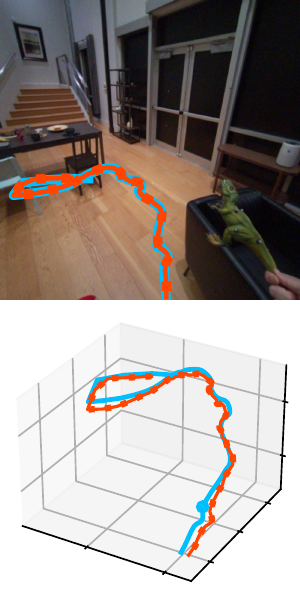}
  \caption{}
\end{subfigure}
\begin{subfigure}{0.24\textwidth}
  \centering
  \includegraphics[clip, height=6.2cm,trim=2cm 0.3cm 2cm 0cm]{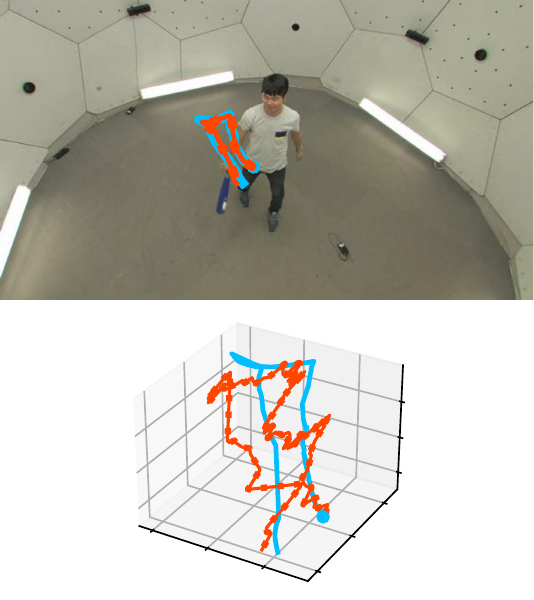}
  \caption{}
\end{subfigure}
\begin{subfigure}{0.24\textwidth}
  \centering
  \includegraphics[clip, height=6.2cm,trim=1.3cm 0.3cm 1.3cm 0cm]{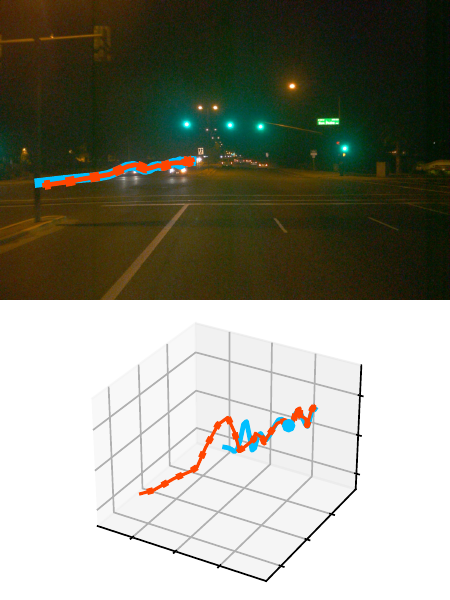}
  \caption{}
\end{subfigure}
\begin{subfigure}{0.24\textwidth}
  \centering
  \includegraphics[clip, height=6.2cm, trim=0cm 0.3cm 0cm 0cm]{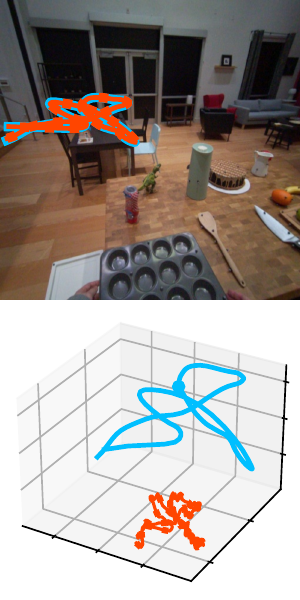}
  \caption{}
\end{subfigure}
\caption{Illustrative TAP-3D results of BootsTAPIR + ZoeDepth. We compare the ground-truth 2D and 3D tracks (blue solid) with the predicted tracks (red dotted). (a) Accurate tracking. (b) Noisy depth estimations result in a noisy 3D track. (c) Inconsistent depth scales across time (scale drift). (d) Inconsistent depth scales across space don't allow a single global scale factor to properly fit all tracks.}
\vspace{-5mm}
\label{fig:failurecases}
\end{figure}

\section{Baselines on TAPVid-3D}
\vspace{-1mm}
\label{sec:baselines}

We construct one set of baselines by combining state-of-art 2D point trackers and monocular depth estimators. In particular, we use for 2D tracking, several state-of-the-art models such as CoTracker ~\cite{karaev2023cotracker}, BootsTAPIR model~\cite{doersch2024bootstap}, and TAPIR~\cite{doersch2023tapir}; and for depth regression, both the monocular depth estimation model ZoeDepth~\cite{bhat2023zoedepth} and the SfM pipeline COLMAP~\cite{schoenberger2016sfm, schoenberger2016mvs}. To convert the frame pixel space predictions into metric $x,y$-position coordinates, we unproject using the camera intrinsics and the $z$-estimate provided by ZoeDepth. In the case of COLMAP, we import the 2D trajectories produced by the TAP methods before running the SfM reconstruction pipeline.

We include the recently released SpatialTracker \cite{xiao2024spatialtracker}, one of the first models designed specifically for 3D point tracking. We also provide results for a new variant of TAPIR~\cite{doersch2023tapir} that we built for 3D tracking, that we call TAPIR-3D (more details in the appendix). Uniquely, this is the only baseline that is trained only on synthetic videos. Finally, we include a static baseline, which projects the pixel query point into 3D, and assumes no motion of that 3D point. Results for all these baseline methods are shown in Table~\ref{table:median_depth_baselines}, for the \texttt{full\_eval} split, and three major methods on \texttt{minival} split. Results in Table~\ref{table:median_depth_baselines} uses median depth scaling; results for other scaling options are included in the appendix.

\begin{table}[h]
    \centering
    \resizebox{\textwidth}{!}{%
    \setlength{\tabcolsep}{3pt}
    \begin{tabular}{rccccccccccccc}
    \toprule
 & \multicolumn{3}{c}{Aria} & \multicolumn{3}{c}{DriveTrack} & \multicolumn{3}{c}{PStudio} & \multicolumn{3}{c}{\textbf{Average}} \\
    Baseline & 3D-AJ $\uparrow$ & APD $\uparrow$ & OA $\uparrow$ & 3D-AJ $\uparrow$ & APD $\uparrow$ & OA $\uparrow$ & 3D-AJ $\uparrow$ & APD $\uparrow$ & OA $\uparrow$
    & 3D-AJ $\uparrow$ & APD $\uparrow$ & OA $\uparrow$ \\
    \cmidrule(r{0.1em}){1-1}\cmidrule(lr{0.1em}){2-4}\cmidrule(lr{0.1em}){5-7}\cmidrule(lr{0.1em}){8-10}\cmidrule(lr{0.1em}){11-13}
Static Baseline & $4.9$ & $10.2$ & $55.4$ & $3.9$ & $6.5$ & $80.8$ & $5.9$ & $11.5$ & $75.8$ & $4.9$ & $9.4$ & $70.7$ \\
TAPIR + COLMAP & $7.1$ & $11.9$ & $72.6$ & $8.9$ & $14.7$ & $80.4$ & $6.1$ & $10.7$ & $75.2$ & $7.4$ & $12.4$ & $76.1$ \\
CoTracker + COLMAP & $8.0$ & $12.3$ & $78.6$ & $11.7$ & $\mathbf{19.1}$ & $81.7$ & $8.1$ & $13.5$ & $77.2$ & $\mathbf{9.3}$ & $15.0$ & $79.1$ \\
BootsTAPIR + COLMAP & $9.1$ & $14.5$ & $78.6$ & $\mathbf{11.8}$ & $18.6$ & $83.8$ & $6.9$ & $11.6$ & $81.8$ & $\mathbf{9.3}$ & $14.9$ & $81.4$ \\
TAPIR + ZoeDepth & $9.0$ & $14.3$ & $79.7$ & $5.2$ & $8.8$ & $81.6$ & $10.7$ & $18.2$ & $78.7$ & $8.3$ & $13.8$ & $80.0$ \\
CoTracker + ZoeDepth & $\mathbf{10.0}$ & $15.9$ & $87.8$ & $5.0$ & $9.1$ & $82.6$ & $11.2$ & $\mathbf{19.4}$ & $80.0$ & $8.7$ & $14.8$ & $83.4$ \\
BootsTAPIR + ZoeDepth & $9.9$ & $\mathbf{16.3}$ & $86.5$ & $5.4$ & $9.2$ & $\mathbf{85.3}$ & $\mathbf{11.3}$ & $19.0$ & $\mathbf{82.7}$ & $8.8$ & $14.8$ & $\mathbf{84.8}$ \\
TAPIR-3D & $2.5$ & $4.8$ & $86.0$ & $3.2$ & $5.9$ & $83.3$ & $3.6$ & $7.0$ & $78.9$ & $3.1$ & $5.9$ & $82.8$ \\
SpatialTracker~\cite{xiao2024spatialtracker} & $9.9$ & $16.1$ & $\mathbf{89.0}$ & $6.2$ & $11.1$ & $83.7$ & $10.9$ & $19.2$ & $78.6$ & $9.0$ & $\mathbf{15.5}$ & $83.7$ \\
\midrule
BootsTAPIR + COLMAP* & $7.3$ & $11.5$ & $76.3$ & $\mathbf{9.3}$ & $\mathbf{15.1}$ & $\mathbf{83.5}$ & $6.2$ & $10.6$ & $78.7$ & $7.6$ & $12.4$ & $79.5$ \\
BootsTAPIR + ZoeDepth* & $8.6$ & $14.5$ & $86.9$ & $5.1$ & $8.7$ & $\mathbf{83.5}$ & $\mathbf{10.2}$ & $\mathbf{17.7}$ & $\mathbf{82.0}$ & $8.0$ & $13.6$ & $\mathbf{84.1}$ \\
SpatialTracker~\cite{xiao2024spatialtracker}* & $\mathbf{9.2}$ & $\mathbf{15.1}$ & $\mathbf{89.9}$ & $5.8$ & $10.2$ & $82.0$ & $9.8$ & $\mathbf{17.7}$ & $78.4$ & $\mathbf{8.3}$ & $\mathbf{14.3}$ & $83.4$ \\
\bottomrule
    \end{tabular}%
    }
\vspace{2mm}
\caption{\textbf{Using median depth scaling}. We compare the performance of 2D-TAP models~\cite{doersch2024bootstap, doersch2023tapir, karaev2023cotracker} combined with ZoeDepth~\cite{bhat2023zoedepth} and COLMAP~\cite{schoenberger2016sfm}. We also include SpatialTracker~\cite{xiao2024spatialtracker}, and a static point baseline. We report the proposed 3D-AJ as well as the APD and OA metrics. The top nine rows are evaluated on the \texttt{full\_eval} split, while the bottom three rows (marked with *) indicate evaluation on the \texttt{minival} split.}
\label{table:median_depth_baselines}
\end{table}

\begin{table}[t]
\footnotesize
    \centering
    \begin{tabular}{rccccccc}
    \toprule
    & Aria & DriveTrack & PStudio & \multicolumn{3}{c}{\textbf{Total}} \\
    & 2D-AJ $\uparrow$ & 2D-AJ $\uparrow$ & 2D-AJ $\uparrow$
    & 2D-AJ $\uparrow$ & APD $\uparrow$ & OA $\uparrow$ \\
    \midrule
    TAPIR~\cite{doersch2023tapir} & $48.6$ & $57.2$ & $48.7$ & $53.2$ & $67.4$ & $80.5$ \\
    CoTracker~\cite{karaev2023cotracker} & $54.2$ & $59.8$ & $51.0$ & $57.2$ & $74.2$ & $84.5$ \\
    BootsTAPIR~\cite{doersch2024bootstap} & $\mathbf{54.7}$ & $\mathbf{62.9}$ & $\mathbf{52.4}$ & $\mathbf{59.1}$ & $\mathbf{74.7}$  & $\mathbf{85.6}$ \\
    \toprule
    \end{tabular}
\vspace{2mm}
\caption{Evaluating the 2D point tracking performance of our baseline models on TAPVid-3D data, by projecting the ground truth 3D trajectories onto the 2D frame.}
\label{table:2dbaselines}
\vspace{-7mm}
\end{table}

Firstly, comparing Tables~\ref{table:median_depth_baselines} and \ref{table:2dbaselines}, we find that the 3D tracking performance of our baselines are significantly lower compared to their 2D tracking abilities. We show examples in Figure~\ref{fig:failurecases}, illustrating common failure modes regressing 3D trajectories, noting that while the 2D trajectories look accurate, their understanding of total scene geometry and correct 3D motion is poor.

Secondly, we find that having the three different data sources provides a video diversity to the benchmark that enables better assessment of the strengths and weakeness of the video model under test. For example, ZoeDepth-based methods perform comparatively well on the PStudio subset, but signficantly underperforms COLMAP on the DriveTrack subset (where monocular depthing might be harder, with the outdoor complex scenes, in which scale-consistent depthing may be difficult). On the other hand, COLMAP struggles with the PStudio subset, where nearly all tracks and majority of the scene is occupied by the main moving objects. This underperformance is likely because COLMAP fails to reconstruct moving content. SpatialTracker uses ZoeDepth under the hood, so its pattern of results across the subsets are similar to 2D Tracking + ZoeDepth. Our TAPIR-3D experimental model, like TAPIR, outputs trajectories independently, so scale consistency across the entire scene/across trajectories is lacking; it is thus correspondingly penalized, and exhibits poor 3D-AJ. The varying strengths and weaknesses of these different methods indicate that the best performance could be achieved by integrating elements from all three.

\textbf{Limitations and Responsible Usage.} The three data sources in our benchmark do not cover all possible domains where users may want to infer dynamics, and automatic annotations may be imperfect, e.g. where the underlying sensor readings have noise (despite our filtering).  Furthermore, we inherit some limitations from TAP and monocular depth: we only evaluate tracking for solid, opaque objects.  From an ethical perspective, Aria and Panoptic were collected in lab settings with consenting participants, while DriveTrack's Waymo videos come from public roads in 6 US cities.  This paper may inherit biases from these datasets, e.g., participants are lab researchers or the populations of those 6 US cities.  This dataset is not intended for training, but care should be taken in training data to avoid biases.  As a benchmark, the broader impacts are similar to those in prior vision and tracking works: there may be very long term applications to activity recognition and surveillance.

\vspace{-2mm}

\section{Conclusion}
\vspace{-1mm}

We introduce TAPVid-3D, a new benchmark for evaluating the nascent Tracking Any Point in 3D (TAP-3D) task. We contributed (1) a pipeline to annotate three distinct real-world video datasets to produce 3D correspondence annotations per video, (2) the first metrics for the TAP-3D task, which measure multiple axis of accuracy of 3D track estimates, and (3) an assessment of the current state of TAP-3D, by evaluating commonly used tracking models.  We believe this benchmark will accelerate research on the TAP-3D task, allowing the development of models with greater dynamic scene understanding from monocular video.

\bibliographystyle{plainnat}
\bibliography{shortstrings, biblio}

\newpage

\begin{center}\textbf{\Large Appendix for TAPVid-3D}\end{center}

\section*{Table of Contents}
\begin{enumerate}
\item \hyperref[sec:samples]{\textcolor{blue}{More Dataset Samples}}
\item \hyperref[sec:stats]{\textcolor{blue}{Dataset Statistics}}
\item \hyperref[sec:metrics1a]{\textcolor{blue}{Metrics using Median, Per-Trajectory, and Local Neighborhood Scaling}}
\item \hyperref[sec:metrics1b]{\textcolor{blue}{Evaluations using Median, Per-Trajectory, and Local Neighborhood Scaling}}
\item \hyperref[sec:metrics2]{\textcolor{blue}{Evaluations using Fixed Metric Distance Thresholds}}
\item \hyperref[sec:baselinesdeets]{\textcolor{blue}{Baselines Details and Compute Resources}}
\item \hyperref[sec:filter]{\textcolor{blue}{Filtering Incorrect Trajectories}}
\item \hyperref[sec:datadeets]{\textcolor{blue}{Dataset Specifications, Metadata, and other Details}}
\item \hyperref[sec:vizsamples]{\textcolor{blue}{Visualized Samples}}
\end{enumerate}

\section{More Dataset Samples}
\label{sec:samples}
More dataset samples are provided on our website, \url{https://tapvid3d.github.io}, including interactive 3D visualizations. 

Finally, we include static visualizations of trajectories in the figures included in the \hyperref[sec:vizsamples]{\textcolor{blue}{Visualized Samples}} section at the end of this PDF. 

\section{Dataset Statistics}
\label{sec:stats}
\begin{figure}[h]
\centering
\begin{subfigure}[b]{0.47\textwidth}
    \includegraphics[width=\textwidth]{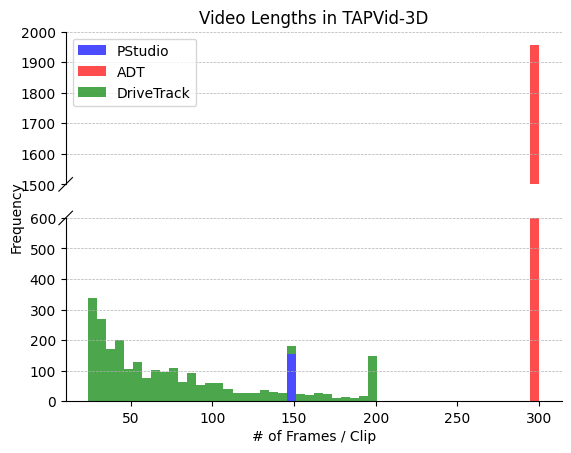}
\end{subfigure}
\hfill
\begin{subfigure}[b]{0.47\textwidth}
    \includegraphics[width=\textwidth]{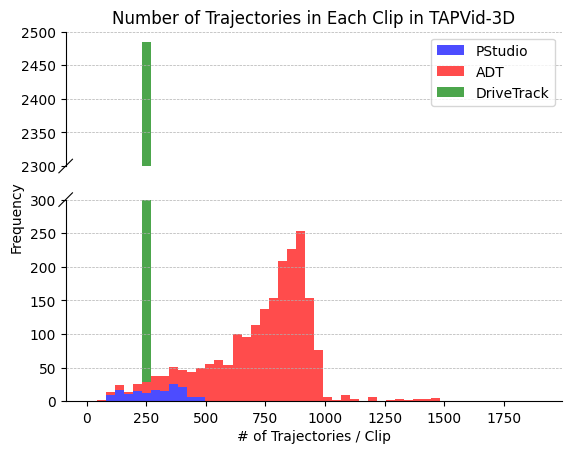}
\end{subfigure}
\bigskip
\begin{subfigure}[b]{0.47\textwidth}
    \includegraphics[width=\textwidth]{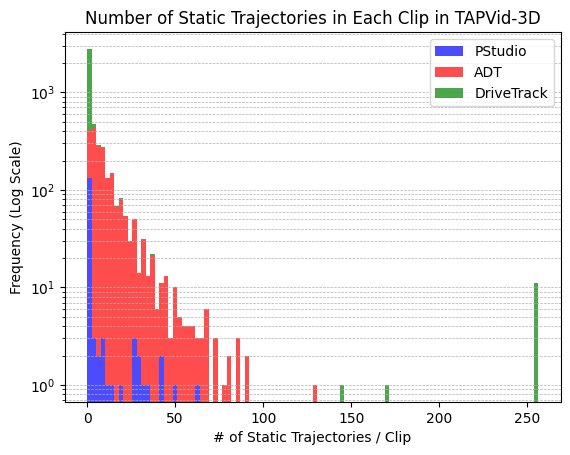}
\end{subfigure}
\hfill
\begin{subfigure}[b]{0.47\textwidth}
    \includegraphics[width=\textwidth]{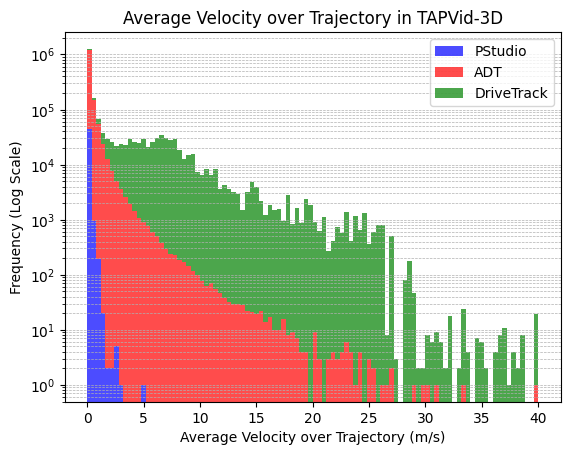}
\end{subfigure}
\caption{Statistics on TAPVid-3D. Top left: video lengths. Top right: Number of trajectories in each clip. Bottom left: Number of static tracks in each clip. Bottom right: average point velocity.}
\label{fig:stats}
\end{figure}

Figure~\ref{fig:stats} showcases various summary statistics about the TAPVid-3D datasets and its 3D point trajectories. In the top left, we have the distribution of the number of frames in each video. The ADT-sourced videos contain the longest videos, and clips of 300 frames were extracted. Similarly Panoptic Studio contains clips of 150 frames, while DriveTrack contains clips of varying duration. In the top right, we have the number of point tracks annotated in each clip. In the bottom right, we count the number of `static' trajectories in each video, marking a trajectory as static if the distance between all pairwise locations within a single point's trajectory is less than 1 centimeter. The roughly 10 DriveTrack videos consisting of static trajectories are usually cars stopped at stoplights. These 'static' videos are a small minority of the 4000+ clips in TAPVid-3D. In the bottom right, we show the average velocity of each trajectory in the dataset, noting that trajectories in DriveTrack are the fastest. These histograms convey that there is a diversity of overall trajectory lengths, video lengths, and point velocities in the TAPVid-3D dataset. Additionally, this dataset is larger than two widely used 2D point tracking real-world-video datasets: TAPVid-Kinetics (1,189 videos) and TAPVid-DAVIS (30 videos).

\section{Metrics using \textit{Median},  \textit{Per-Trajectory}, and \textit{Local Neighborhood} Rescaling}
\label{sec:metrics1a}
In the results included in the main paper, we compute the 3D Average Jaccard and APD metrics using a \emph{global median} rescaling procedure (L277).  To get a good score, the entire scene must be reconstructed up to scale, and dynamic objects must be placed precisely.  This is useful for many applications, such as navigation, but for others it may be overly stringent.  If there is little camera motion, or if some of the objects have unclear size, it may also be very difficult for models to infer global scene shape.

However, not all applications require such strong global scene shape capabilities.  For example, for imitation learning, we may want an agent to simply approach an object. For such applications, measuring the relative depth of estimated 3D locations along a \textit{single trajectory} may be sufficient, and it may be substantially easier, as the (2D) scaling of local textures may provide enough information to solve the problem.  More generally, for robotic imitation of an assembly task, it is the \textit{local} consistency that's most important: as long as points that are near each other in 3D have the correct  depth relative to one another, then the relative pose of the assembled parts will be clear, especially at the critical stage when the assembled parts are close together.  

To enable more rapid progress in such domains, we propose two additional approaches to rescaling estimated trajectories to match the ground-truth 3D point cloud:  \textit{Per-Trajectory}, and \textit{Local Neighborhood}.  We apply the same Average Jaccard metric regardless of how the points are rescaled, although in the case of Local Neighborhood, the ground truth trajectories are also slightly modified, as explained below. In all cases, users can evaluate the same predictions using any metric without providing any extra information.

Per-Trajectory scaling is computed by rescaling each track $P^i$ separately, multiplying by $\|P^i_{t_q}\|/\|\hat{P}^i_{t_q}\|$, where $t_q$ is the query frame index, and then computing 3D AJ as before.  As a result, methods must only compute the \textit{relative} depth for the point at each time, relative to the query frame.

Local Neighborhood is somewhat more involved.  Here, the goal is to capture whether \textit{nearby} points are scaled correctly relative to one another, even if distant parts of the scene may not be.  For example, if the goal is to understand an action depicted in a video that uses a tool, it is typically important to understand where the hand is relative to the tool, and where the tool is relative to the objects it's acting on.  The distance to the backgrounds--such as the back wall of the room--may not be obvious, especially if there's relatively little camera motion.  However, precisely computing these distances is not relevant to understanding the tool's motion.

Intuitively, we wish to find an intermediate between two extremes: either rescaling the entire scene with a single scale factor, or rescaling every point with its own scale factor.  
To this end, we propose to scale each track according to the other track segments that intersect with its \textit{4D tubelet}~\cite{liu2023centertube}, according to a fixed neighborhood radius.

Specifically, we start by choosing a single neighborhood radius $\tau$ specified in meters.  For a given trajectory $P^i$,  we first find all points that are within $\tau$ meters of the ground truth on any frame, which define the tubelet $\mathcal{T}(P^i)$ associated to the trajectory $P^i$:

\[
\mathcal{T}(P^i) = \{P^j_t \hspace{.5em}s.t.\hspace{.5em} \|P^j_t-P^i_t\|<\tau\}.
\]

Note that tubelet $\mathcal{T}(P^i)$ includes the trajectory $P^i$ entirely, plus the portions of the trajectories of the other points where they come closer than the tublet's radius $\tau$. For each selected ground-truth point in the tubelet, we select the corresponding points from the predictions to construct $\mathcal{T}(P^i; \hat{P}^i)$ as

\[
\mathcal{T}(P^i; \hat{P}^i) = \{\hat{P}^j_t \hspace{.5em}s.t.\hspace{.5em} \|P^j_t-P^i_t\|<\tau\}.
\]

Analogously, we select the predicted and ground-truth visibility $\mathcal{T}(\hat{v}^i)$ and $\mathcal{T}(v^i)$.

Finally, given a predicted tubelet $\mathcal{T}(P^i; \hat{P}^i)$, we rescale all its points together using the ratio of query point distances $\|P^i_{t_q}\|/\|\hat{P}^i_{t_q}\|$, and evaluate the rescaled tubelet set as if it were a single trajectory, by replacing $P^i$, $\hat{P}^i$ and $v_i$ with their tubelet counterparts in equations (5) or (6) to compute the $APD_{3D}$ and $AJ_{3D}$ respectively. In our experiments, we set the radius $\tau$ to 3 centimeters for the PStudio and Aria scenes, and as 10 centimeters for the DriveTrack scenes, across all experiments. This is because tabletop manipulation and human-object motion likely require finer-grained movement than large vehicle movement in the Waymo Open public road scenes.

\begin{table}[h]
    \centering
    \resizebox{\textwidth}{!}{%
    \setlength{\tabcolsep}{3pt}
    \begin{tabular}{rccccccccccccc}
    \toprule
 & \multicolumn{3}{c}{Aria} & \multicolumn{3}{c}{DriveTrack} & \multicolumn{3}{c}{PStudio} & \multicolumn{3}{c}{\textbf{Average}} \\
    Baseline & 3D-AJ $\uparrow$ & APD $\uparrow$ & OA $\uparrow$ & 3D-AJ $\uparrow$ & APD $\uparrow$ & OA $\uparrow$ & 3D-AJ $\uparrow$ & APD $\uparrow$ & OA $\uparrow$
    & 3D-AJ $\uparrow$ & APD $\uparrow$ & OA $\uparrow$ \\
    \cmidrule(r{0.1em}){1-1}\cmidrule(lr{0.1em}){2-4}\cmidrule(lr{0.1em}){5-7}\cmidrule(lr{0.1em}){8-10}\cmidrule(lr{0.1em}){11-13}
Static Baseline & $5.4$ & $11.8$ & $55.4$ & $4.8$ & $8.4$ & $80.8$ & $6.4$ & $12.7$ & $75.8$ & $5.5$ & $11.0$ & $70.7$ \\
TAPIR + COLMAP & $26.5$ & $37.7$ & $72.6$ & $16.5$ & $24.6$ & $80.4$ & $12.1$ & $19.6$ & $75.2$ & $18.4$ & $27.3$ & $76.1$ \\
CoTracker + COLMAP & $26.8$ & $38.3$ & $78.6$ & $18.2$ & $28.8$ & $81.7$ & $12.1$ & $19.7$ & $77.2$ & $19.1$ & $28.9$ & $79.1$ \\
BootsTAPIR + COLMAP & $\mathbf{28.8}$ & $\mathbf{41.3}$ & $78.6$ & $\mathbf{20.0}$ & $\mathbf{29.3}$ & $83.8$ & $12.9$ & $20.8$ & $81.8$ & $\mathbf{20.6}$ & $\mathbf{30.4}$ & $81.4$ \\
TAPIR + ZoeDepth & $16.2$ & $24.2$ & $79.7$ & $7.4$ & $12.2$ & $81.6$ & $12.0$ & $20.0$ & $78.7$ & $11.9$ & $18.8$ & $80.0$ \\
CoTracker + ZoeDepth & $17.4$ & $26.3$ & $87.8$ & $6.7$ & $12.3$ & $82.6$ & $12.0$ & $20.8$ & $80.0$ & $12.0$ & $19.8$ & $83.4$ \\
BootsTAPIR + ZoeDepth & $17.3$ & $27.0$ & $86.5$ & $7.4$ & $12.3$ & $\mathbf{85.3}$ & $12.3$ & $20.6$ & $\mathbf{82.7}$ & $12.3$ & $20.0$ & $\mathbf{84.8}$ \\
TAPIR-3D & $8.5$ & $14.9$ & $86.0$ & $10.2$ & $17.0$ & $83.3$ & $7.2$ & $13.1$ & $78.9$ & $8.6$ & $15.0$ & $82.8$ \\
SpatialTracker~\cite{xiao2024spatialtracker} & $17.4$ & $26.9$ & $\mathbf{89.0}$ & $9.0$ & $16.1$ & $83.7$ & $\mathbf{14.2}$ & $\mathbf{24.6}$ & $78.6$ & $13.6$ & $22.5$ & $83.7$ \\
    \bottomrule
    \end{tabular}%
    }
\vspace{2mm}
\caption{\textbf{Using per-trajectory depth scaling}. We compare the performance on the \texttt{full\_eval} split of several 2D-TAP models~\cite{doersch2024bootstap, doersch2023tapir, karaev2023cotracker} combined with ZoeDepth~\cite{bhat2023zoedepth} and COLMAP~\cite{schoenberger2016sfm} on the TAPVid-3D benchmark. We also measure performance on the recently released SpatialTracker~\cite{xiao2024spatialtracker}, and a static point baseline, in which the predicted trajectories are exactly the same as the query point.}
\label{table:per_trajectory_baselines}
\end{table}

\begin{table}[h]
    \centering
    \resizebox{\textwidth}{!}{%
    \setlength{\tabcolsep}{3pt}
    \begin{tabular}{rccccccccccccc}
    \toprule
 & \multicolumn{3}{c}{Aria} & \multicolumn{3}{c}{DriveTrack} & \multicolumn{3}{c}{PStudio} & \multicolumn{3}{c}{\textbf{Average}} \\
    Baseline & 3D-AJ $\uparrow$ & APD $\uparrow$ & OA $\uparrow$ & 3D-AJ $\uparrow$ & APD $\uparrow$ & OA $\uparrow$ & 3D-AJ $\uparrow$ & APD $\uparrow$ & OA $\uparrow$
    & 3D-AJ $\uparrow$ & APD $\uparrow$ & OA $\uparrow$ \\
    \cmidrule(r{0.1em}){1-1}\cmidrule(lr{0.1em}){2-4}\cmidrule(lr{0.1em}){5-7}\cmidrule(lr{0.1em}){8-10}\cmidrule(lr{0.1em}){11-13}
Static Baseline & $5.5$ & $11.8$ & $56.0$ & $4.8$ & $8.4$ & $80.8$ & $6.4$ & $12.6$ & $75.7$ & $5.5$ & $10.9$ & $70.8$ \\
TAPIR + COLMAP & $23.8$ & $34.4$ & $72.8$ & $12.9$ & $20.3$ & $80.4$ & $9.9$ & $16.5$ & $75.1$ & $15.5$ & $23.7$ & $76.1$ \\
CoTracker + COLMAP & $24.6$ & $35.3$ & $78.8$ & $15.7$ & $\mathbf{25.2}$ & $81.7$ & $10.8$ & $17.7$ & $77.1$ & $17.0$ & $26.1$ & $79.2$ \\
BootsTAPIR + COLMAP & $\mathbf{26.1}$ & $\mathbf{38.0}$ & $78.8$ & $\mathbf{16.6}$ & $25.1$ & $83.8$ & $10.8$ & $17.8$ & $81.8$ & $\mathbf{17.8}$ & $\mathbf{27.0}$ & $81.5$ \\
TAPIR + ZoeDepth & $15.7$ & $23.5$ & $79.8$ & $6.3$ & $10.5$ & $81.6$ & $11.2$ & $18.9$ & $78.7$ & $11.0$ & $17.6$ & $80.1$ \\
CoTracker + ZoeDepth & $17.0$ & $25.7$ & $88.0$ & $6.0$ & $10.9$ & $82.6$ & $11.4$ & $19.9$ & $80.0$ & $11.4$ & $18.8$ & $83.5$ \\
BootsTAPIR + ZoeDepth & $16.8$ & $26.3$ & $86.7$ & $6.4$ & $10.9$ & $\mathbf{85.3}$ & $11.6$ & $19.6$ & $\mathbf{82.6}$ & $11.6$ & $18.9$ & $\mathbf{84.9}$ \\
TAPIR-3D & $7.3$ & $12.9$ & $86.3$ & $5.9$ & $10.5$ & $83.4$ & $5.1$ & $9.6$ & $78.9$ & $6.1$ & $11.0$ & $82.8$ \\
SpatialTracker~\cite{xiao2024spatialtracker} & $16.7$ & $25.7$ & $\mathbf{89.3}$ & $6.9$ & $12.4$ & $83.7$ & $\mathbf{12.3}$ & $\mathbf{21.6}$ & $78.5$ & $12.0$ & $19.9$ & $83.8$ \\
    \bottomrule
    \end{tabular}%
    }
    \vspace{2mm}
\caption{\textbf{Using local neighborhood scaling}. We compare the performance on the \texttt{full\_eval} split of 2D-TAP models~\cite{doersch2024bootstap, doersch2023tapir, karaev2023cotracker} combined with ZoeDepth~\cite{bhat2023zoedepth} and COLMAP~\cite{schoenberger2016sfm}. We also include SpatialTracker~\cite{xiao2024spatialtracker}, and a static point baseline.}
\label{table:local_neighborhood_baselines}
\end{table}

\section{Evaluations with Median, Per-Trajectory, and Local Neighborhood Scaling}
\label{sec:metrics1b}
Tables~\ref{table:per_trajectory_baselines}, \ref{table:local_neighborhood_baselines}, \ref{table:median_depth_baselines} present additional experimental results on the \texttt{full\_eval} set, on all our baselines. To avoid biasing the results to the TAPVid-3D splits with higher number of videos, these tables present averaged results across the three constituent data sources (weighing each source equally). 

As expected, the AJ increases when using the less-strict local rescaling approaches. That is, \emph{per-trajectory} scaling require less scale consistency than the \emph{local neighborhood} metric, which itself is less stringent than the \emph{global median scaling}.  However, different methods improve by different amounts.  Perhaps most surprisingly, COLMAP gives strong performance with local and per-trajectory rescaling, but underperforms ZoeDepth on Aria and Panoptic Studio when evaluated with global rescaling.  This is likely because COLMAP completely fails to reconstruct moving content.  For scenes where the majority of tracks are moving, the median rescaling will fail completely; therefore, ZoeDepth giving reasonable estimates for a larger fraction of points gives it an advantage.  

TAPIR-3D, unsurprisingly, presents poor performance using global or local scaling, as it does not provide relative depth estimates for different tracks.  However, evaluated with per-trajectory scaling, it gives competitive results on DriveTrack, even outperforming SpaTracker.  This is somewhat surprising given that it operates on a completely different principle than other methods; it is trained using entirely synthetic data and does not use any geometric constraints, nor relies on monodepth models providing geometric priors.  Overall, the different strengths and weaknesses of these highly-diverse methods suggests that the best performance will come from a method that combines ideas from all three.  SpaTracker is a step in this direction, using a monodepth initialization while checking the 2D consistency of 3D reconstructions via reprojection, similar to COLMAP, and it often gives competitive performance.  We hope that this benchmark can provide a way to quantify how well future methods in this vein accomplish the task.

Finally, we include a static baseline, which predicts the static 3D point $P_q = K^{-1}[x_q, y_q, 1]^T\cdot Z_q$ for all timestamps, in order to quantify the impact of motion.  Note that this baseline still requires ground truth depth for the query points, and so isn't trivial to reproduce automatically; however, it still performs very poorly, even for the PStudio dataset where the camera is static (note that the static baseline is static \emph{in the camera coordinate frame}).  Thus, we conclude tracking the camera and tracking the objects is important for obtaining strong performance.

\section{Evaluations using Fixed Metric Distance Thresholds}
\label{sec:metrics2}
In the Average Jaccard formulation in Section 3.5, we describe how we use determine correctly predicted points along a trajectory, using a depth-adaptive radius threshold denoted $\delta_{3D}(P_t^i)$. We also explored using a fixed metric threshold. Specifically, instead of the $\{1, 2, 4, 8, 16\}$ pixel thresholds (projected into 3D), we use a the fixed metric radius thresholds of 1 centimeter, 4 centimeter, 16 centimeters, 64 centimeters, and 2.56 meters. If the predicted point is within this distance to the ground truth point, it is marked as position correct within that threshold. Table~\ref{table:fixed_metric_baselines} describes the model baselines results using this alternative metric.

\begin{table}[h]
    \centering
    \resizebox{\textwidth}{!}{%
    \setlength{\tabcolsep}{3pt}
    \begin{tabular}{rccccccccccccc}
    \toprule
 & \multicolumn{3}{c}{Aria} & \multicolumn{3}{c}{DriveTrack} & \multicolumn{3}{c}{PStudio} & \multicolumn{3}{c}{\textbf{Average}} \\
    Baseline & 3D-AJ $\uparrow$ & APD $\uparrow$ & OA $\uparrow$ & 3D-AJ $\uparrow$ & APD $\uparrow$ & OA $\uparrow$ & 3D-AJ $\uparrow$ & APD $\uparrow$ & OA $\uparrow$
    & 3D-AJ $\uparrow$ & APD $\uparrow$ & OA $\uparrow$ \\
    \cmidrule(r{0.1em}){1-1}\cmidrule(lr{0.1em}){2-4}\cmidrule(lr{0.1em}){5-7}\cmidrule(lr{0.1em}){8-10}\cmidrule(lr{0.1em}){11-13}
Static Baseline & $21.2$ & $40.3$ & $55.4$ & $5.8$ & $9.7$ & $80.8$ & $31.7$ & $46.4$ & $75.8$ & $19.6$ & $32.1$ & $70.7$ \\
TAPIR + COLMAP & $20.5$ & $32.2$ & $72.6$ & $15.5$ & $21.9$ & $80.4$ & $28.8$ & $41.7$ & $75.2$ & $21.6$ & $31.9$ & $76.1$ \\
CoTracker + COLMAP & $23.6$ & $33.6$ & $78.6$ & $18.2$ & $\mathbf{25.7}$ & $81.7$ & $31.6$ & $44.1$ & $77.2$ & $24.4$ & $34.4$ & $79.1$ \\
BootsTAPIR + COLMAP & $24.0$ & $35.5$ & $78.6$ & $\mathbf{18.7}$ & $25.2$ & $83.8$ & $31.6$ & $43.6$ & $81.8$ & $24.7$ & $34.7$ & $81.4$ \\
TAPIR + ZoeDepth & $28.2$ & $41.7$ & $79.7$ & $11.0$ & $15.9$ & $81.6$ & $39.0$ & $55.2$ & $78.7$ & $26.1$ & $37.6$ & $80.0$ \\
CoTracker + ZoeDepth & $32.7$ & $44.0$ & $87.8$ & $10.7$ & $16.2$ & $82.6$ & $40.2$ & $\mathbf{56.1}$ & $80.0$ & $27.8$ & $38.8$ & $83.4$ \\
BootsTAPIR + ZoeDepth & $31.9$ & $\mathbf{45.6}$ & $86.5$ & $11.4$ & $16.3$ & $\mathbf{85.3}$ & $\mathbf{41.5}$ & $55.9$ & $\mathbf{82.7}$ & $28.3$ & $39.2$ & $84.8$ \\
TAPIR-3D & $19.2$ & $29.9$ & $86.0$ & $7.0$ & $10.9$ & $83.3$ & $24.8$ & $36.1$ & $78.9$ & $17.0$ & $25.6$ & $82.8$ \\
SpatialTracker~\cite{xiao2024spatialtracker} & $\mathbf{33.1}$ & $45.0$ & $\mathbf{89.0}$ & $13.1$ & $19.3$ & $83.7$ & $39.5$ & $\mathbf{56.1}$ & $78.6$ & $\mathbf{28.5}$ & $\mathbf{40.2}$ & $\mathbf{83.7}$ \\
    \bottomrule
    \end{tabular}%
    }
\vspace{2mm}
\caption{\textbf{Using fixed metric thresholds for AJ, with median scaling}. We compare the performance on the \texttt{full\_eval} split of several 2D-TAP models~\cite{doersch2024bootstap, doersch2023tapir, karaev2023cotracker} combined with ZoeDepth~\cite{bhat2023zoedepth} and COLMAP~\cite{schoenberger2016sfm}. We also include SpatialTracker~\cite{xiao2024spatialtracker} and the static point baseline.}
\label{table:fixed_metric_baselines}
\end{table}

\section{Baselines Details and Compute Resources}
\label{sec:baselinesdeets}
\textbf{CoTracker}. We use the pretrained model and PyTorch code from the official CoTracker codebase and run inference enabling the bi-directional tracking mode, with no other modifications to the default parameters. Internally, inference is performed in $512\times 384$ resolution, and the output predictions are rescaled back to the original clip resolutions. Inference is performed using A100 GPUs, and processing each dataset clip takes about 30s, totaling roughly 38 GPUh for running CoTracker on the whole benchmark.

\textbf{BootsTAPIR and TAPIR}. We use the pretrained models and JAX code from the official codebase and run inference with the default parameters. Internally, inference is performed in $256\times 256$ resolution, and the output predictions are rescaled back to the original clip resolutions. Inference was performed in a CPU cluster using up to 1024 CPUs and totalling about 22800 CPUh.

\textbf{COLMAP}. For running COLMAP, we dumped the 2D tracks estimated by CoTracker, TAPIR and BootsTAPIR as a set of per-frame image feature files and corresponding matches in txt format and imported those in COLMAP using the `feature\_importer` and `matches\_importer` functionality. We then perform 3D reconstruction through the incremental mapping pipeline (`mapper`). As each input 2D track can lead to multiple reconstructed 3D points across time (eg. for moving objects), we only keep those with larger "track length" (number of images where that 3D point was reconstructed from). Finally, we transform the resulting reconstructed 3D points positions in world coordinates to camera coordinates using the predicted extrinsic parameters. Inference was performed in a CPU cluster using up to 1024 CPUs and totalling about 14000 CPUh.

\textbf{ZoeDepth}. We used the pretrained models and PyTorch code from the official codebase and run inference with the default parameters. Inference is performed in the native resolution for the ADT and Panoptic Studio clips and in $720\times 480$ for DriveTrack, where the original resolution was too large for running infrence in this model on a standard GPU. Inference was performed on 16 V100 GPUs, totalling about 200GPUh.

\subsection*{TAPIR3D}

We propose a straightforward extension of TAPIR to 3D by training on 3D ground truth from Kubric~\cite{greff2022kubric}.  As Kubric is synthetic data, it is straightforward to obtain ground-truth 3D point tracks.  However, we don't expect that Monodepth models trained on Kubric will generalize, as the scenes are very different from real ones.  However, we expect that a model trained here might be able to estimate \textit{relative} depth of a point at different times on the trajectory, relative to the query point.  Like with point tracking, we expect that low-level texture information may be sufficient to predict the relative depth (specifically, the scaling of the texture elements), and so high-level semantic understanding won't be necessary, meaning that it can be learned from a semantically meaningless dataset.

Specifically, we train TAPIR to output the log depth of each point on the trajectory, relative to the query point.  This is a scalar quantity, and can be predicted using the same network structure as the other scalar quantites, e.g., the occlusion logit.  That is, we predict an initial log-space scale factor for every frame by adding an extra head to TAPIR's occlusion prediction network, which performs convolutions on top of the cost volume for each frame followed by global average pooling.  Then we feed this estimate to the iterative refinement steps by concatenating it with the local score maps, the initial occlusion estimate, and so on, passing it through the 1-D convolutional network which produces an update; again, we add an extra head on this network which produces and updated relative depth estimate.  We apply an L1 loss on the estimate for both the initialization and each of the four refinement passes, with a weight of 1.0.  Otherwise, we train the entire network using the procedure described in~\cite{doersch2023tapir}.

\section{Filtering Incorrect Trajectories}
\label{sec:filter}
We apply different automatic filters for removing problematic tracks. Tracks can present three type of issues: (i) issues with visibility flags, (ii) queries which are outside the moving objects, and (iii) noisy 3D trajectories. 

We found visibility issues (i) to be present in all dataset splits, and we remove it simply by oversampling the number of query points and discarding those whose visibility flag changes state more than a 10\% of the number of frames in the video.

Issue (ii) was present mostly in the DriveTrack split, where trajectories in a video are localized and describe the motion of exactly one moving object in the scene. In some cases the 3D point-clouds associated with vehicles also contain points that are within the object bounding box, but outside of the object itself, such as in the road. To filter out errant trajectories, we use the Segment Anything model (SAM) to generate an object mask for each frame~\cite{kirillov2023segment}. We prompt SAM with a point prompt, computed by taking the geometric median of DriveTrack trajectories at each point in time. 

Finally, we found that noisy 3D trajectories (iii) could occur in the  Panoptic Studio split, where sometimes the reconstructed 3D Gaussians where not sufficiently constrained due to surfaces having uniform colors. In this case we apply a similar approach as before, and score trajectories based on the percentage of time they are on  foreground object masks across all camera viewpoints. We perform a hyperparameter search on the threshold value and select the points that stay on the object masks at least 75\% of the time across all masks, which removes most of the problematic points.

\section{Dataset Specifications, Metadata, and other Details}
\label{sec:datadeets}
For this dataset release, we preserve the licenses for the constituent original data sources, which are non-commercial. For our additions, to the extent that we can, we release under a standard Apache 2.0 license.  A full amalgamated license will be available in the open-sourced repository during complete release of the work, after the review period is finished.

We will publicly host the dataset for wide consumption by researchers on Google Cloud Storage indefinitely. Part of the dataset is already hosted in this way (and how the Colab link linked above is able to run). We also intend to open-source code for computing the new 3D-AJ metrics after the camera ready. We anticipate the release will require little maintenance (and the TAPVid-2D dataset release that the team released two years ago is similarly low maintanence), but we are happy to address any emergent issues raised by users.

Specific implementation details on how the dataset can be read are found in the Colab link provided. Each dataset example is provided in a \texttt{*.npy} file, containing the fields: \texttt{tracks\_xyz} of shape [T, Q, 3] (containing the Q ground truth point tracks for the corresponding video of T frames, with $(x, y, z)$-coordinates in meters), \texttt{query\_xyt} of shape [Q, 3] (containing each track's query point position in format (x, y, t), in (x,y)-pixel space and $t$ as the query frame index), the ground truth \texttt{visibility} flags with shape [Q, T], and the \texttt{camera\_intrinsics} (as $[f_x, f_y, c_x, c_y]$). Each \texttt{*.npy} file is named after its corresponding video in the original data source, which can be loaded by downloading from the original hosting sites~\cite{sun2020waymoopen,pan2023aria,balasingam2023drivetrack}, respecting their corresponding licenses.

\section{Visualized Samples}
\label{sec:vizsamples}

\begin{figure}[h]
    \centering
    \setlength{\tabcolsep}{2pt}
    \renewcommand{\arraystretch}{1}
    \begin{tabular}{ccc}
        \includegraphics[width=0.32\linewidth]{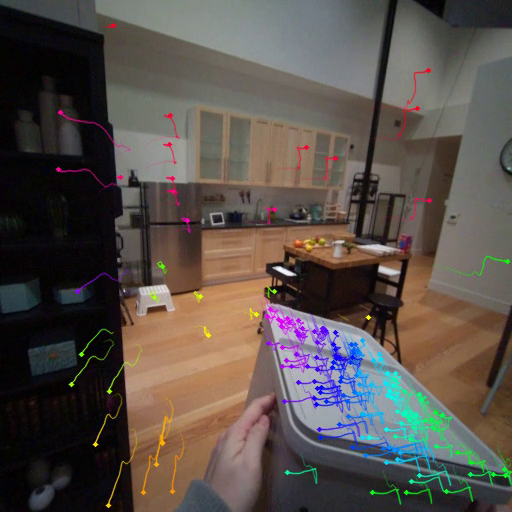} & 
        \includegraphics[width=0.32\linewidth]{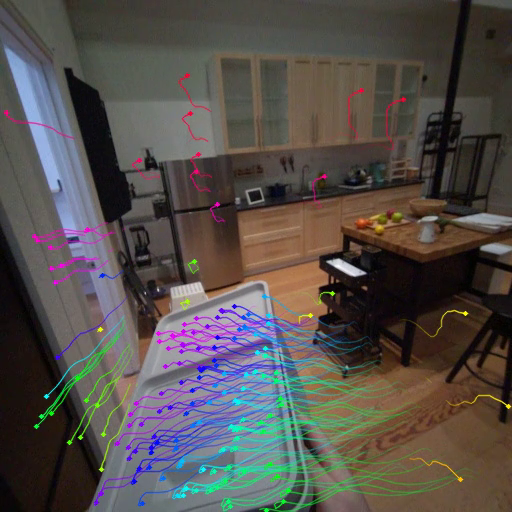} & 
        \includegraphics[width=0.32\linewidth]{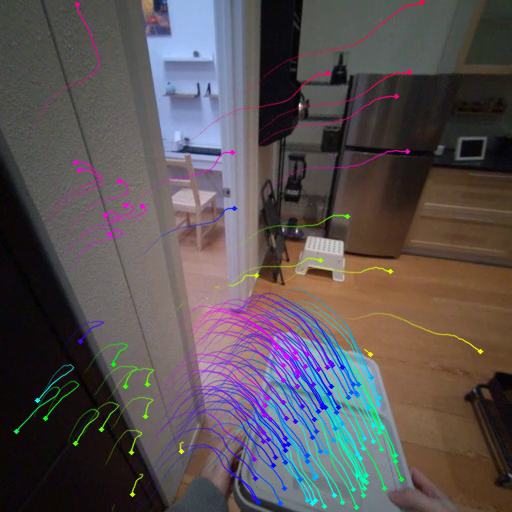} \\
        
        \includegraphics[clip, height=0.28\linewidth, trim=2.5cm 2cm 2.5cm 2cm]{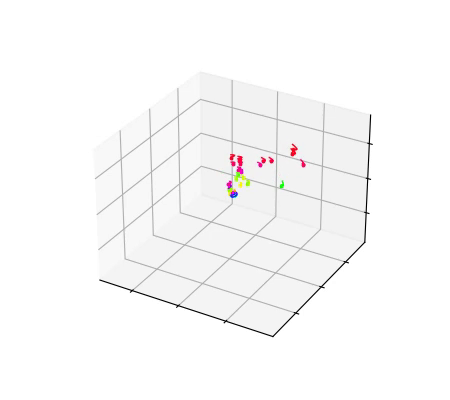} & 
        \includegraphics[clip, height=0.28\linewidth, trim=2.5cm 2cm 2.5cm 2cm]{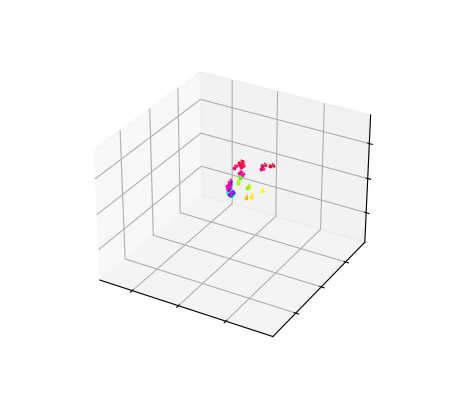} & 
        \includegraphics[clip, height=0.28\linewidth, trim=2.5cm 2cm 2.5cm 2cm]{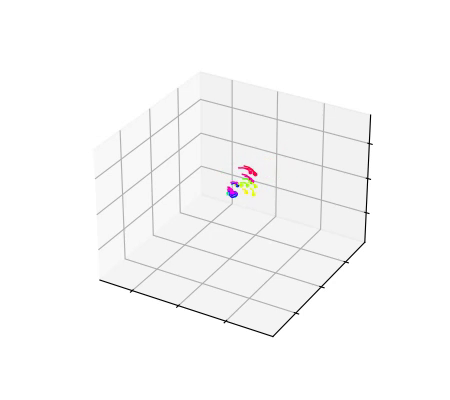} \\ \\

        \includegraphics[width=0.32\linewidth]{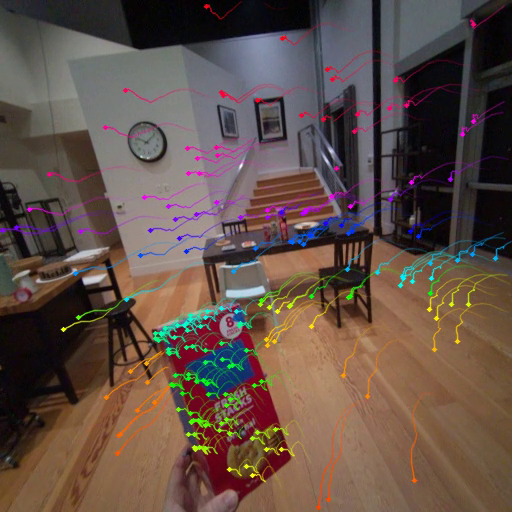} & 
        \includegraphics[width=0.32\linewidth]{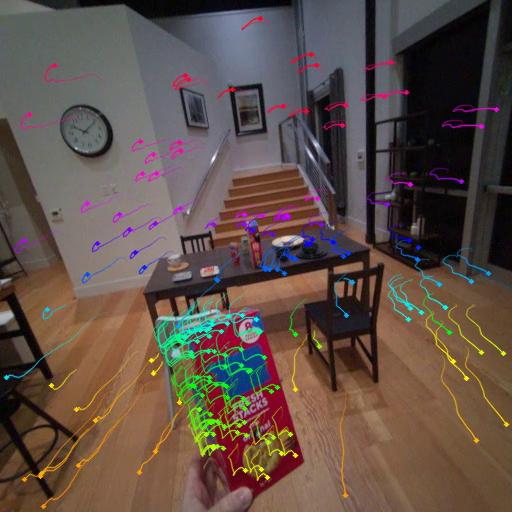} & 
        \includegraphics[width=0.32\linewidth]{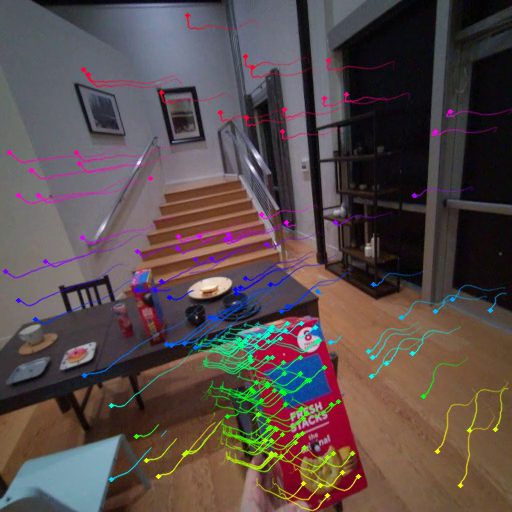} \\
        
        \includegraphics[clip, height=0.28\linewidth, trim=2.5cm 2cm 2.5cm 2cm]{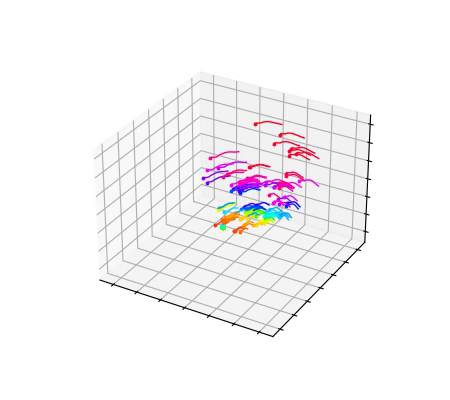} & 
        \includegraphics[clip, height=0.28\linewidth, trim=2.5cm 2cm 2.5cm 2cm]{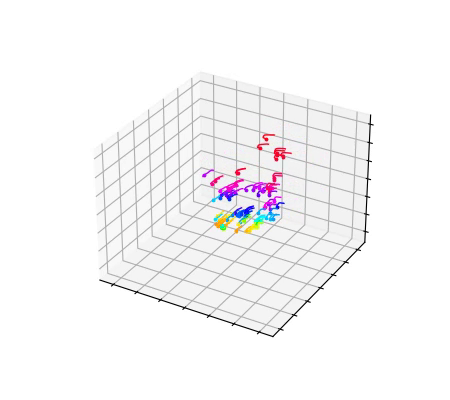} & 
        \includegraphics[clip, height=0.28\linewidth, trim=2.5cm 2cm 2.5cm 2cm]{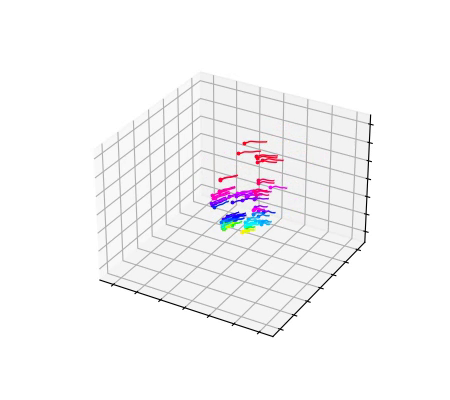} \\
    \end{tabular}
    \caption{Random samples from ADT subset in TAPVid-3D: on the top row, we visualize the point trajectories projected into the 2D video frame; on the bottom row, we visualize the metric 3D point trajectories. For each video, we show 3 frames sampled at time step 30, 60 and 90.}
    \label{fig:samples1}
\end{figure}

\begin{figure}[htbp]
    \centering
    \resizebox{\textwidth}{!}{%
    \setlength{\tabcolsep}{2pt}
    \renewcommand{\arraystretch}{1}
    \begin{tabular}{ccc}
    
        \includegraphics[width=0.32\linewidth]{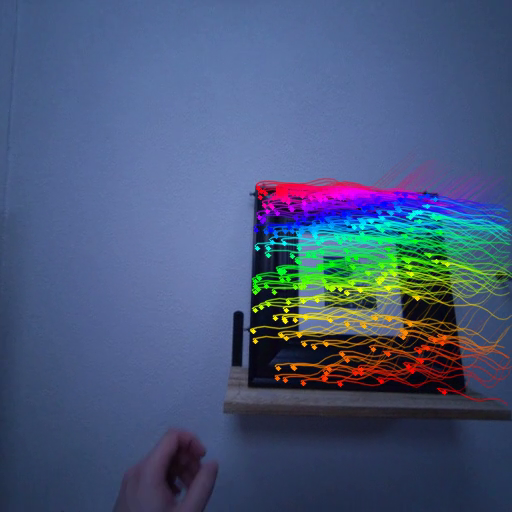} & 
        \includegraphics[width=0.32\linewidth]{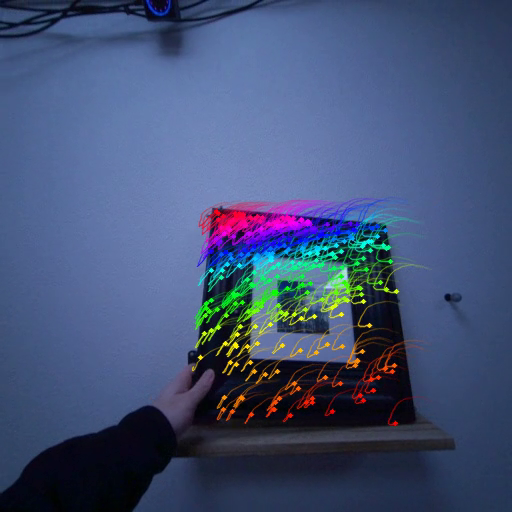} & 
        \includegraphics[width=0.32\linewidth]{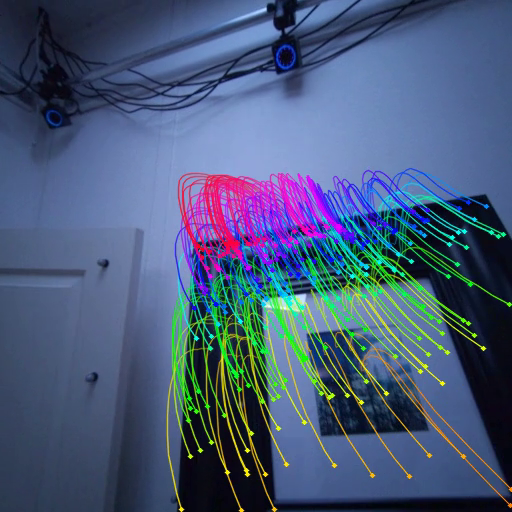} \\
        
        \includegraphics[clip, height=0.28\linewidth, trim=2.5cm 2cm 2.5cm 2cm]{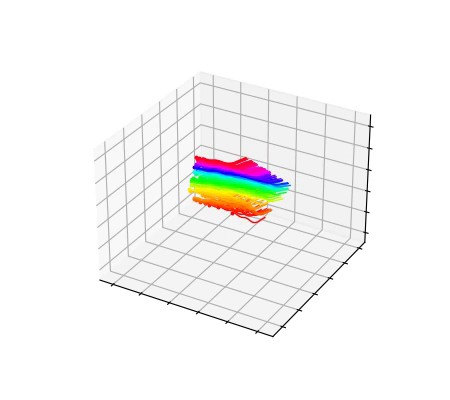} & 
        \includegraphics[clip, height=0.28\linewidth, trim=2.5cm 2cm 2.5cm 2cm]{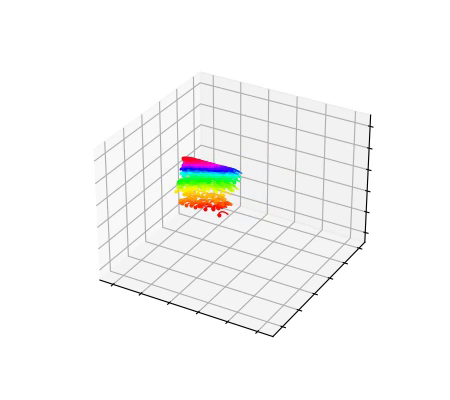} & 
        \includegraphics[clip, height=0.28\linewidth, trim=2.5cm 2cm 2.5cm 2cm]{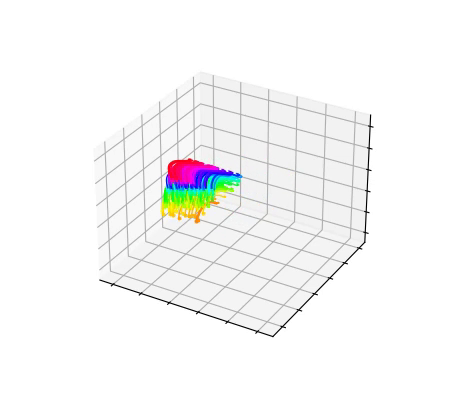} \\ \\
        
        \includegraphics[width=0.32\linewidth]{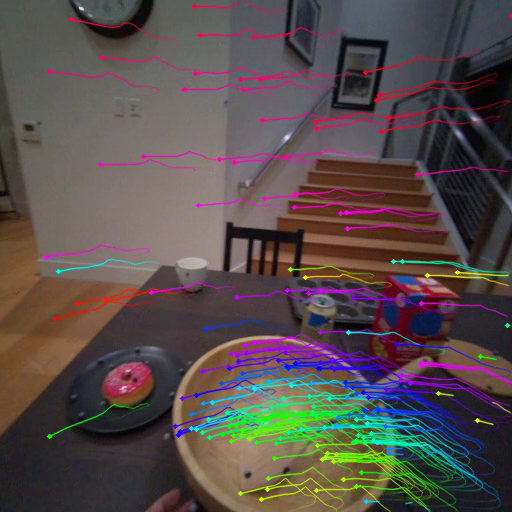} & 
        \includegraphics[width=0.32\linewidth]{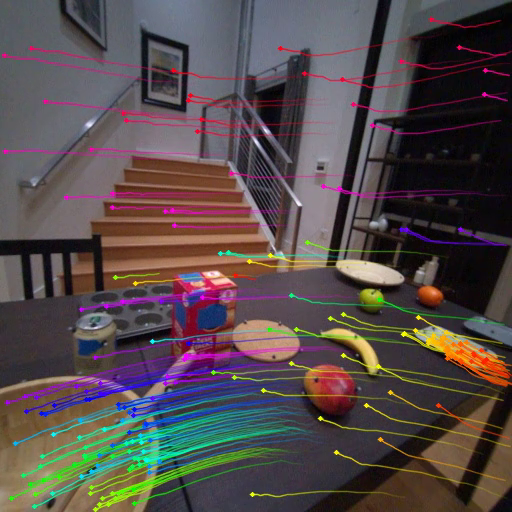} & 
        \includegraphics[width=0.32\linewidth]{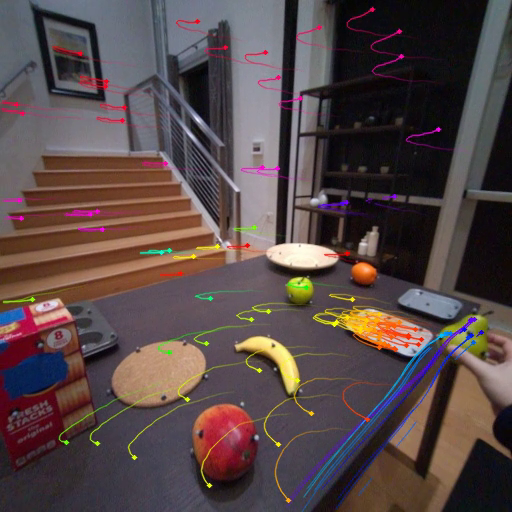} \\
        
        \includegraphics[clip, height=0.28\linewidth, trim=2.5cm 2cm 2.5cm 2cm]{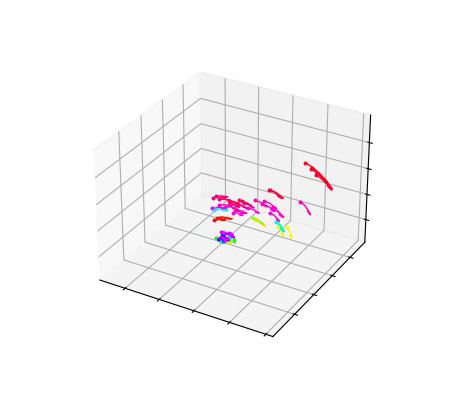} & 
        \includegraphics[clip, height=0.28\linewidth, trim=2.5cm 2cm 2.5cm 2cm]{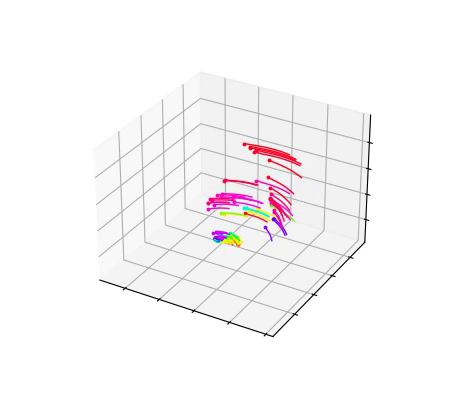} & 
        \includegraphics[clip, height=0.28\linewidth, trim=2.5cm 2cm 2.5cm 2cm]{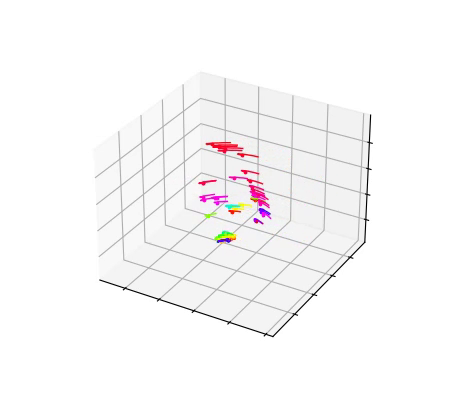} \\

    \end{tabular}
    }
    \caption{Random samples from ADT subset in TAPVid-3D (cont'd.): on the top row, we visualize the point trajectories projected into the 2D video frame; on the bottom row, we visualize the metric 3D point trajectories. For each video, we show 3 frames sampled at time step 30, 60 and 90.}
    \label{fig:samples2}
\end{figure}

\begin{figure}[htbp]
    \centering
    \resizebox{\textwidth}{!}{%
    \setlength{\tabcolsep}{2pt}
    \renewcommand{\arraystretch}{1}
    \begin{tabular}{ccc}
        \includegraphics[width=0.32\linewidth]{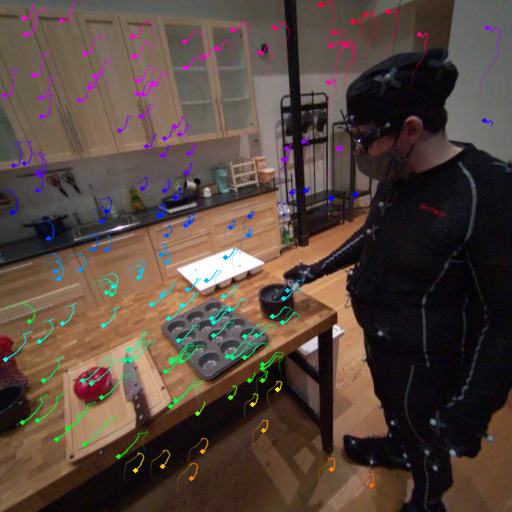} & 
        \includegraphics[width=0.32\linewidth]{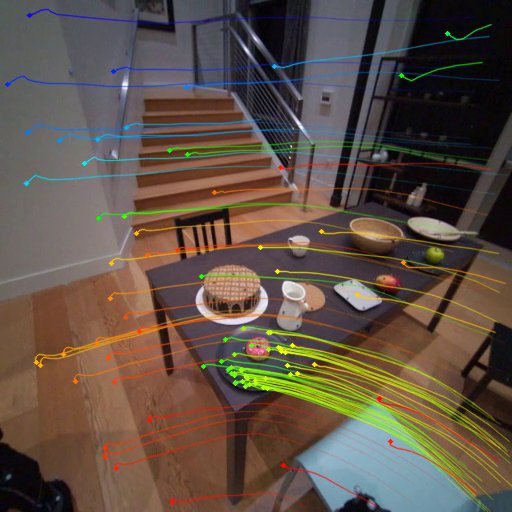} & 
        \includegraphics[width=0.32\linewidth]{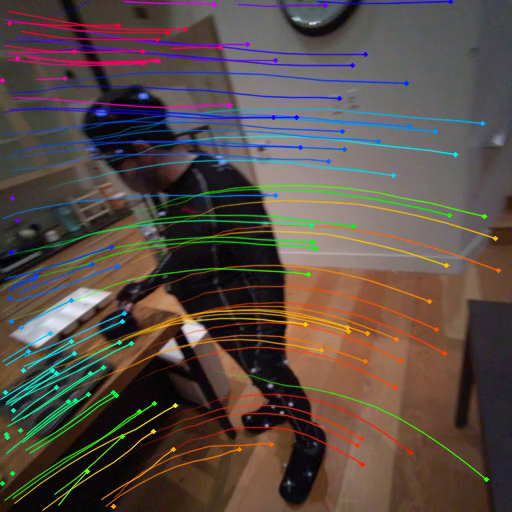} \\
    
        \includegraphics[clip, height=0.28\linewidth, trim=2.5cm 2cm 2.5cm 2cm]{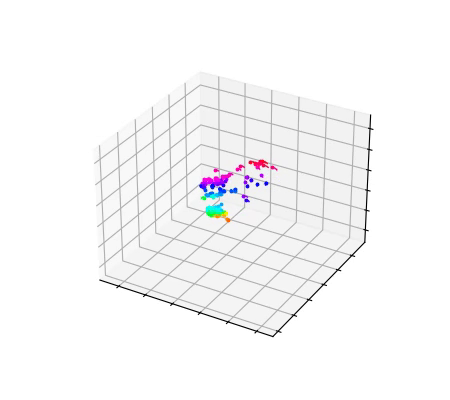} & 
        \includegraphics[clip, height=0.28\linewidth, trim=2.5cm 2cm 2.5cm 2cm]{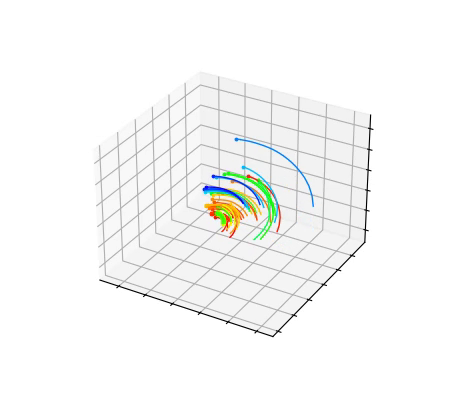} & 
        \includegraphics[clip, height=0.28\linewidth, trim=2.5cm 2cm 2.5cm 2cm]{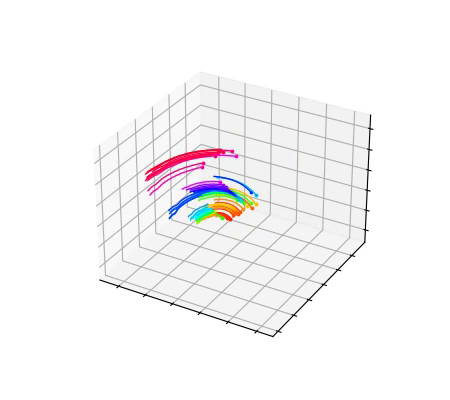} \\ \\
        
        \includegraphics[width=0.32\linewidth]{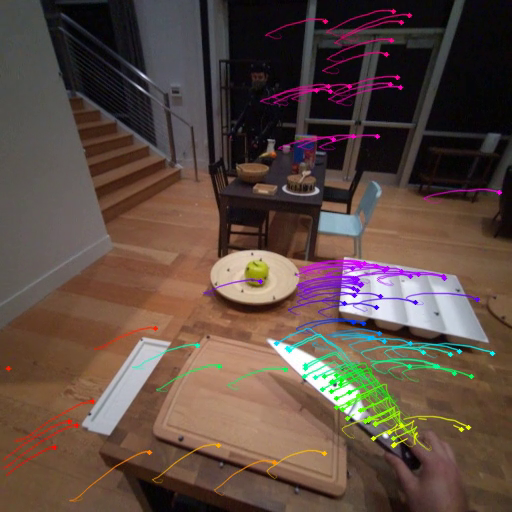} & 
        \includegraphics[width=0.32\linewidth]{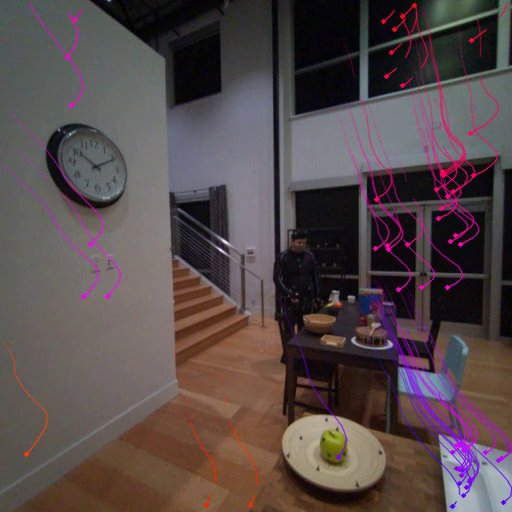} & 
        \includegraphics[width=0.32\linewidth]{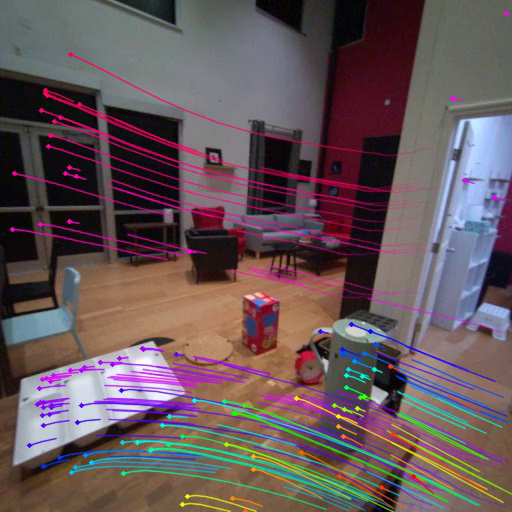} \\
        
        \includegraphics[clip, height=0.28\linewidth, trim=2.5cm 2cm 2.5cm 2cm]{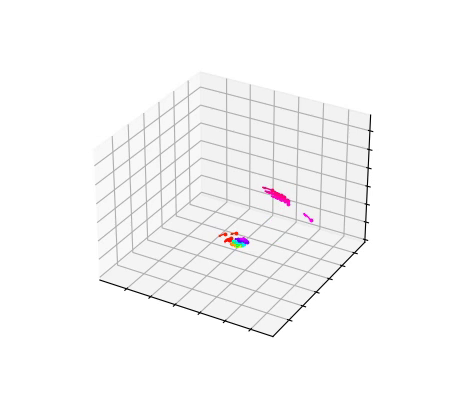} & 
        \includegraphics[clip, height=0.28\linewidth, trim=2.5cm 2cm 2.5cm 2cm]{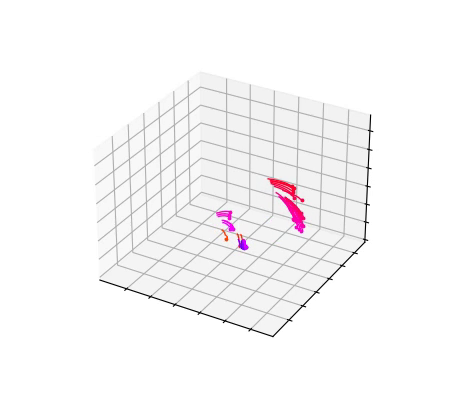} & 
        \includegraphics[clip, height=0.28\linewidth, trim=2.5cm 2cm 2.5cm 2cm]{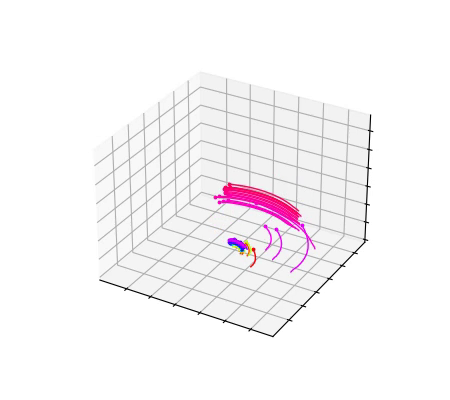} \\
    \end{tabular}
    }
    \caption{Random samples from ADT subset in TAPVid-3D (cont'd.): on the top row, we visualize the point trajectories projected into the 2D video frame; on the bottom row, we visualize the metric 3D point trajectories. For each video, we show 3 frames sampled at time step 30, 60 and 90.}
    \label{fig:samples3}
\end{figure}

\begin{figure}[h]
    \centering
    \setlength{\tabcolsep}{2pt}
    \renewcommand{\arraystretch}{1}
    \begin{tabular}{ccc}
        \includegraphics[width=0.32\linewidth]{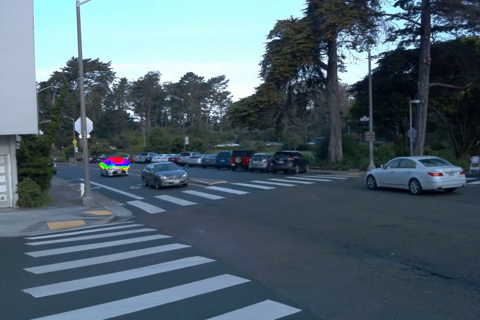} & 
        \includegraphics[width=0.32\linewidth]{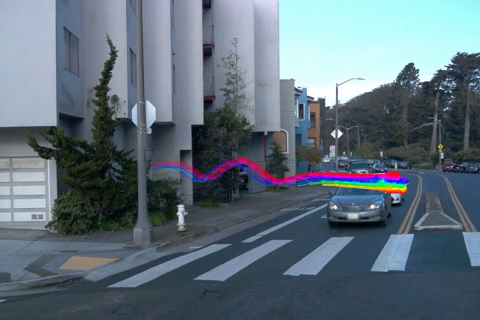} & 
        \includegraphics[width=0.32\linewidth]{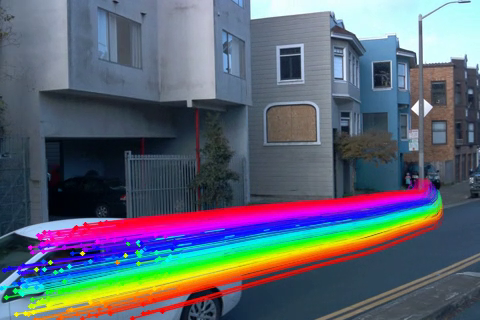} \\
        
        \includegraphics[clip, height=0.28\linewidth, trim=2.5cm 2cm 2.5cm 2cm]{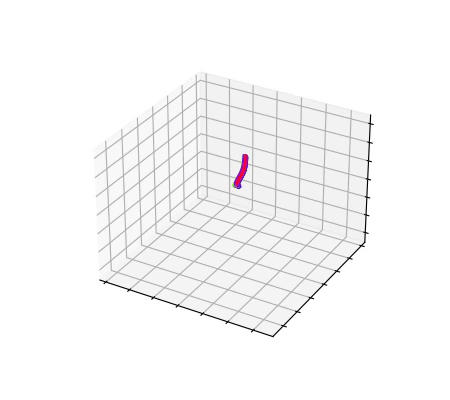} & 
        \includegraphics[clip, height=0.28\linewidth, trim=2.5cm 2cm 2.5cm 2cm]{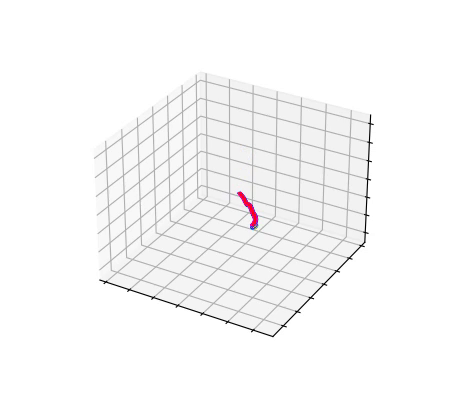} & 
        \includegraphics[clip, height=0.28\linewidth, trim=2.5cm 2cm 2.5cm 2cm]{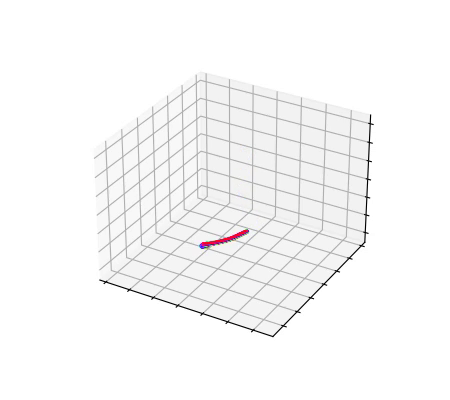} \\ \\

        \includegraphics[width=0.32\linewidth]{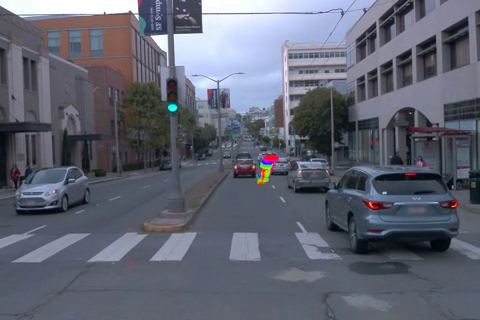} & 
        \includegraphics[width=0.32\linewidth]{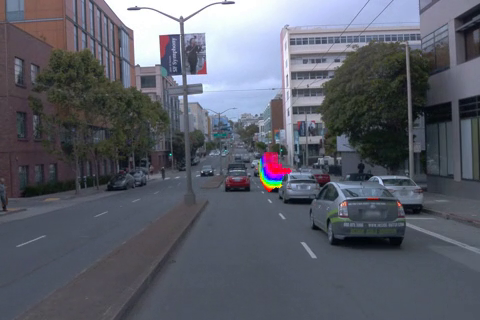} & 
        \includegraphics[width=0.32\linewidth]{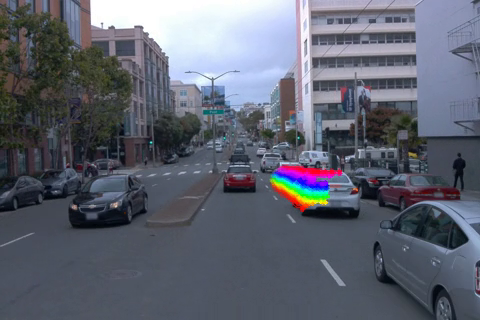} \\
        
        \includegraphics[clip, height=0.28\linewidth, trim=2.5cm 2cm 2.5cm 2cm]{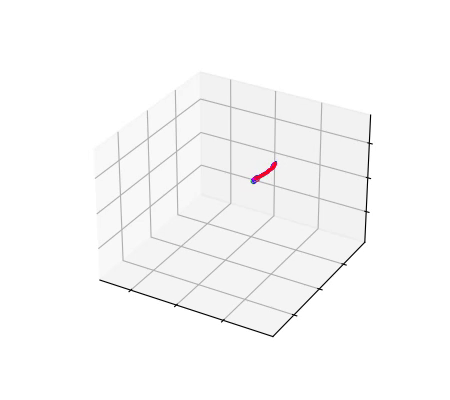} & 
        \includegraphics[clip, height=0.28\linewidth, trim=2.5cm 2cm 2.5cm 2cm]{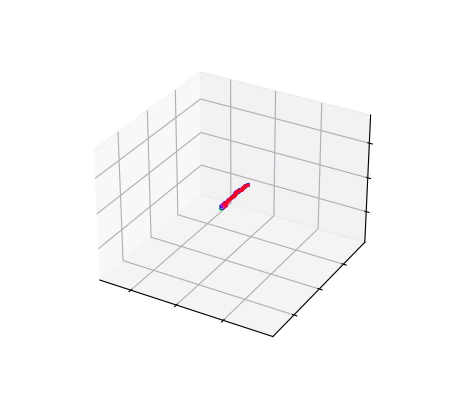} & 
        \includegraphics[clip, height=0.28\linewidth, trim=2.5cm 2cm 2.5cm 2cm]{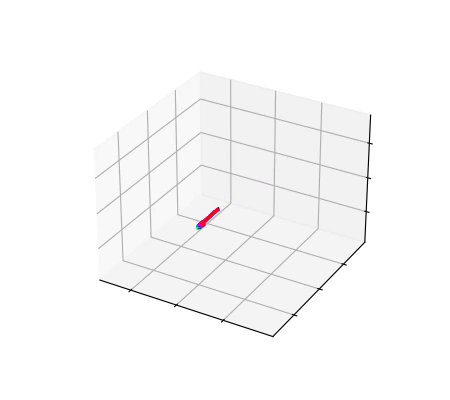} \\
    \end{tabular}
    \caption{Random samples from DriveTrack subset in TAPVid-3D: on the top row, we visualize the point trajectories projected into the 2D video frame; on the bottom row, we visualize the metric 3D point trajectories. For each video, we show 3 frames sampled at time step 30, 60 and 90.}
    \label{fig:samples4}
\end{figure}

See Figures~\ref{fig:samples1}, \ref{fig:samples2}, \ref{fig:samples3}, \ref{fig:samples4}, \ref{fig:samples5}, \ref{fig:samples6}, \ref{fig:samples7}, \ref{fig:samples8}, and \ref{fig:samples9} below.

\begin{figure}[htbp]
    \centering
    \resizebox{\textwidth}{!}{%
    \setlength{\tabcolsep}{2pt}
    \renewcommand{\arraystretch}{1}
    \begin{tabular}{ccc}
    
        \includegraphics[width=0.32\linewidth]{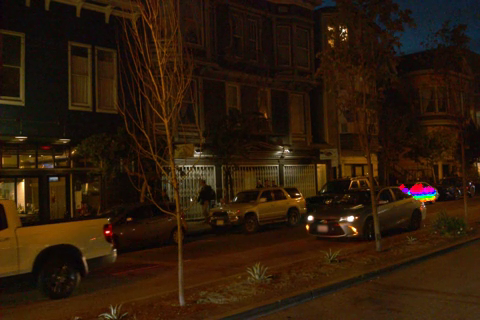} & 
        \includegraphics[width=0.32\linewidth]{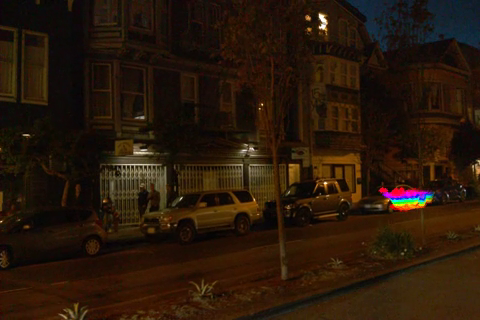} & 
        \includegraphics[width=0.32\linewidth]{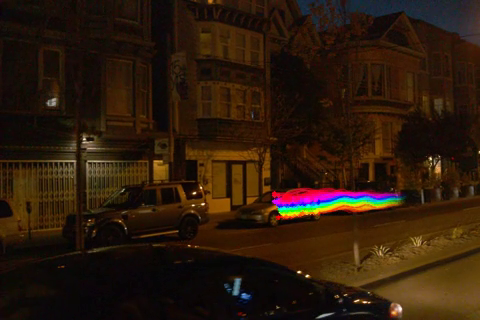} \\
        
        \includegraphics[clip, height=0.28\linewidth, trim=2.5cm 2cm 2.5cm 2cm]{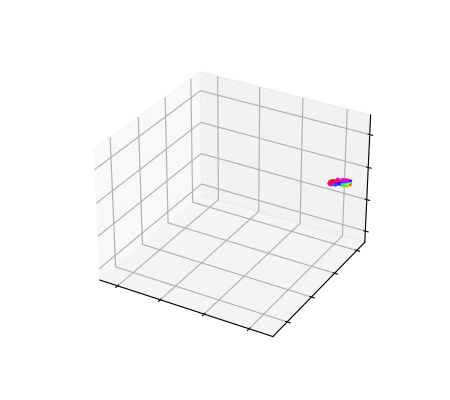} & 
        \includegraphics[clip, height=0.28\linewidth, trim=2.5cm 2cm 2.5cm 2cm]{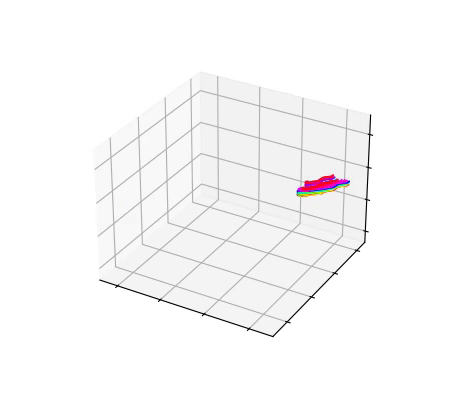} & 
        \includegraphics[clip, height=0.28\linewidth, trim=2.5cm 2cm 2.5cm 2cm]{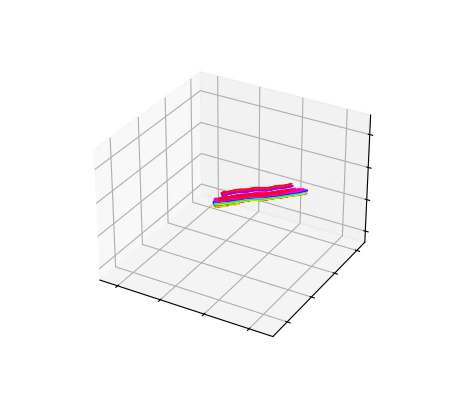} \\ \\
        
        \includegraphics[width=0.32\linewidth]{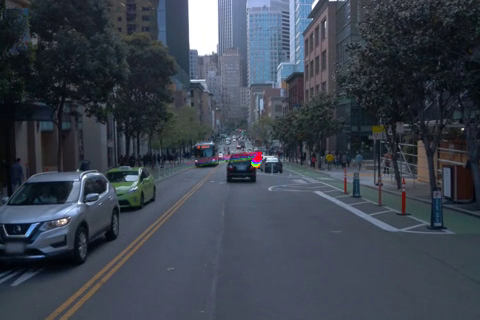} & 
        \includegraphics[width=0.32\linewidth]{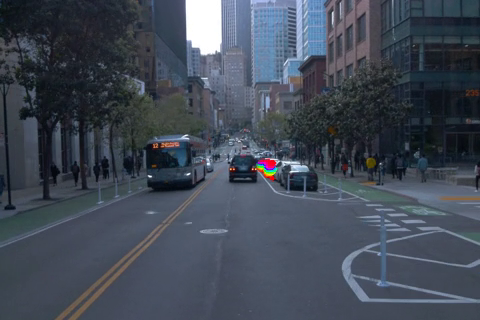} & 
        \includegraphics[width=0.32\linewidth]{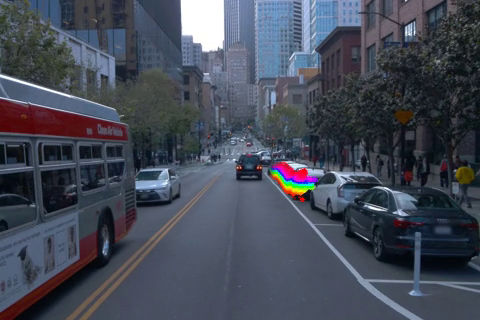} \\ 
        
        \includegraphics[clip, height=0.28\linewidth, trim=2.5cm 2cm 2.5cm 2cm]{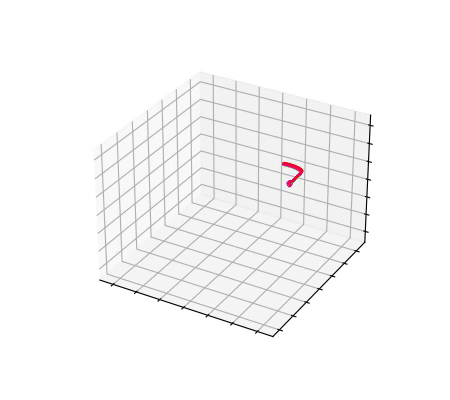} & 
        \includegraphics[clip, height=0.28\linewidth, trim=2.5cm 2cm 2.5cm 2cm]{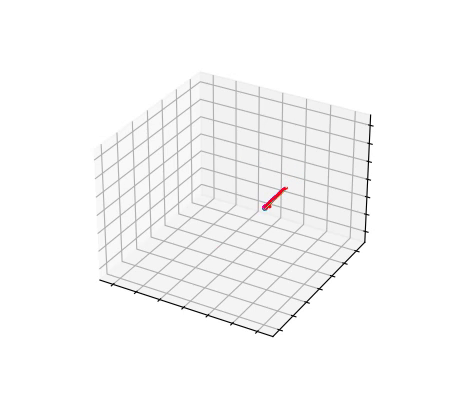} & 
        \includegraphics[clip, height=0.28\linewidth, trim=2.5cm 2cm 2.5cm 2cm]{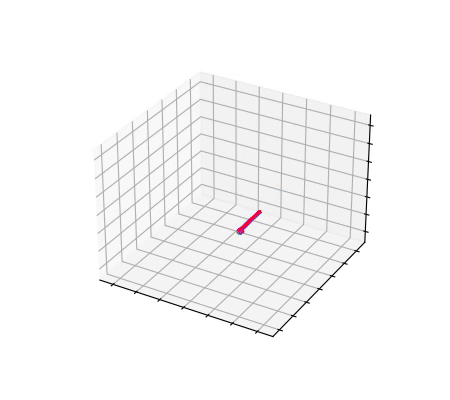} \\
    \end{tabular}
    }
    \caption{Random samples from DriveTrack subset in TAPVid-3D (cont'd.): on the top row, we visualize the point trajectories projected into the 2D video frame; on the bottom row, we visualize the metric 3D point trajectories. For each video, we show 3 frames sampled at time step 30, 60 and 90.}
    \label{fig:samples5}
\end{figure}

\begin{figure}[htbp]
    \centering
    \resizebox{\textwidth}{!}{%
    \setlength{\tabcolsep}{2pt}
    \renewcommand{\arraystretch}{1}
    \begin{tabular}{ccc}
        \includegraphics[width=0.32\linewidth]{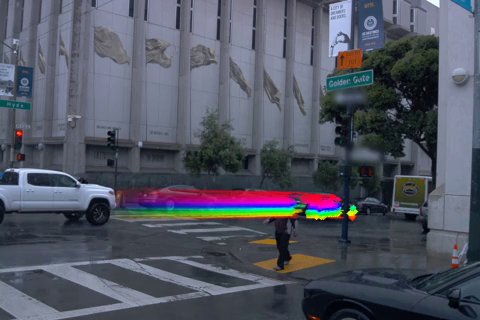} & 
        \includegraphics[width=0.32\linewidth]{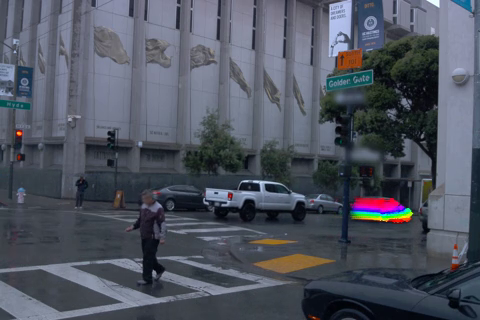} & 
        \includegraphics[width=0.32\linewidth]{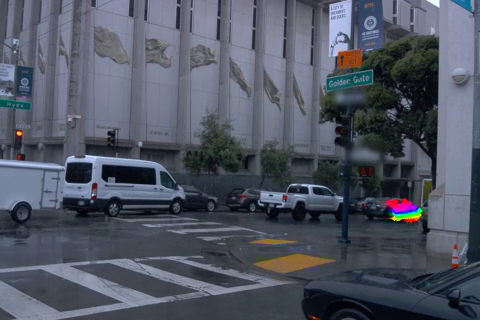} \\
    
        \includegraphics[clip, height=0.28\linewidth, trim=2.5cm 2cm 2.5cm 2cm]{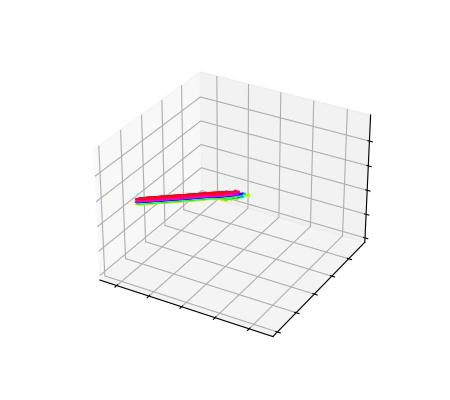} & 
        \includegraphics[clip, height=0.28\linewidth, trim=2.5cm 2cm 2.5cm 2cm]{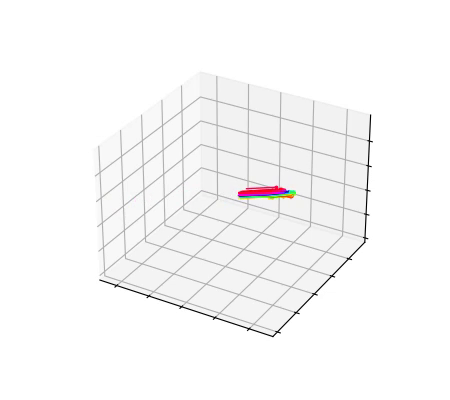} & 
        \includegraphics[clip, height=0.28\linewidth, trim=2.5cm 2cm 2.5cm 2cm]{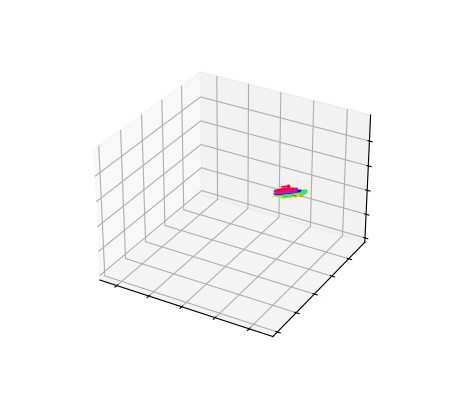} \\ \\
        
        \includegraphics[width=0.32\linewidth]{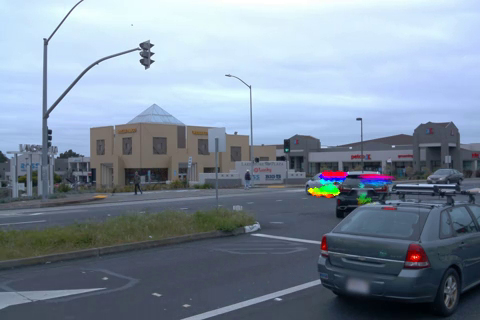} & 
        \includegraphics[width=0.32\linewidth]{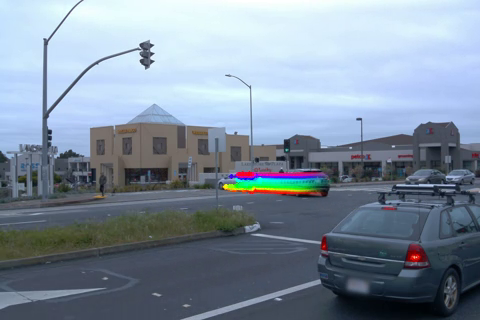} & 
        \includegraphics[width=0.32\linewidth]{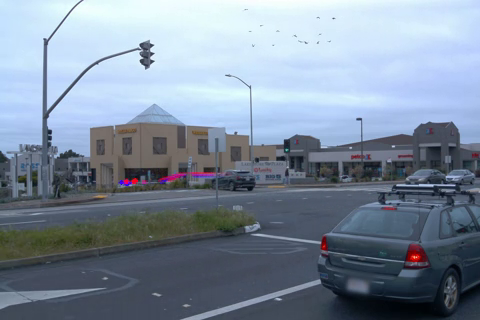} \\ 
        
        \includegraphics[clip, height=0.28\linewidth, trim=2.5cm 2cm 2.5cm 2cm]{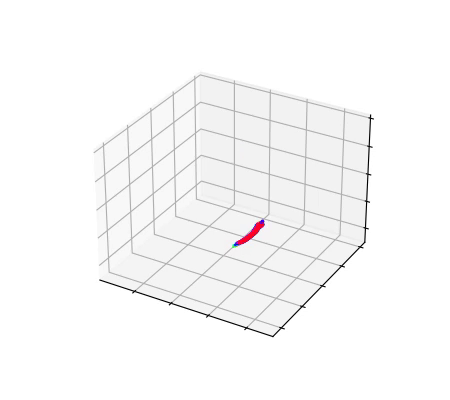} & 
        \includegraphics[clip, height=0.28\linewidth, trim=2.5cm 2cm 2.5cm 2cm]{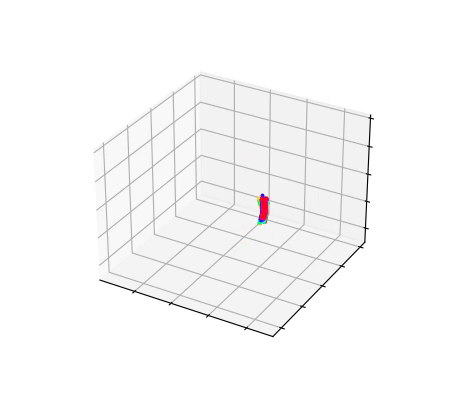} & 
        \includegraphics[clip, height=0.28\linewidth, trim=2.5cm 2cm 2.5cm 2cm]{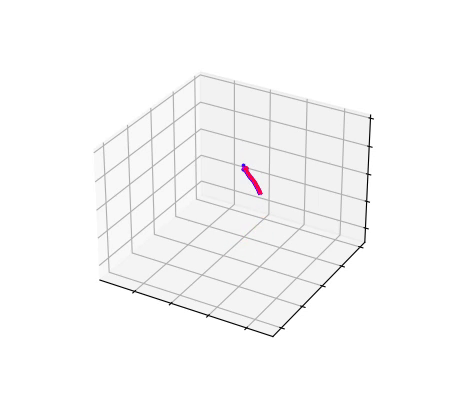} \\
    \end{tabular}
    }
    \caption{Random samples from DriveTrack subset in TAPVid-3D (cont'd.): on the top row, we visualize the point trajectories projected into the 2D video frame; on the bottom row, we visualize the metric 3D point trajectories. For each video, we show 3 frames sampled at time step 30, 60 and 90.}
    \label{fig:samples6}
\end{figure}

\begin{figure}[h]
    \centering
    \setlength{\tabcolsep}{2pt}
    \renewcommand{\arraystretch}{1}
    \begin{tabular}{ccc}
        \includegraphics[width=0.32\linewidth]{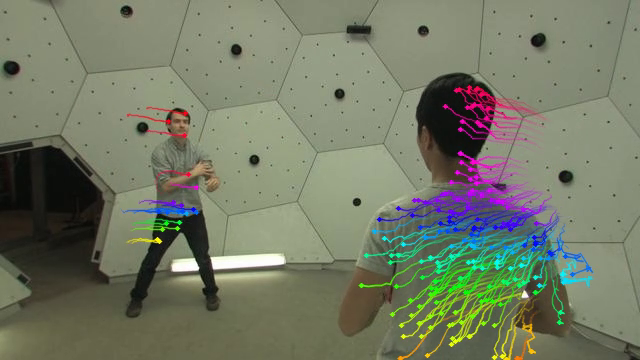} & 
        \includegraphics[width=0.32\linewidth]{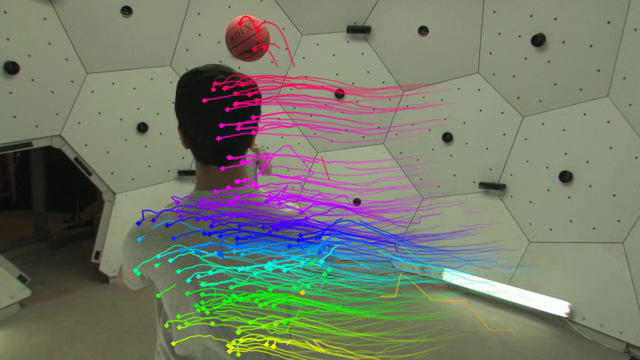} & 
        \includegraphics[width=0.32\linewidth]{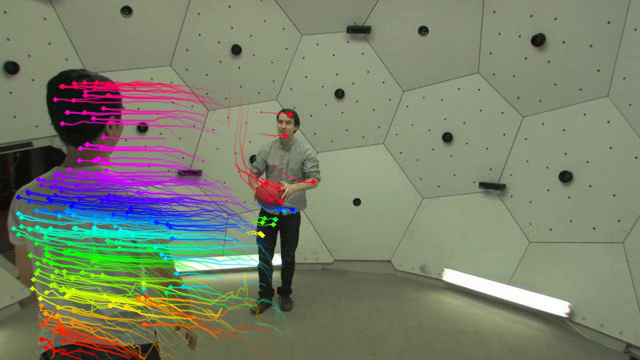} \\
        
        \includegraphics[clip, height=0.28\linewidth, trim=2.5cm 2cm 2.5cm 2cm]{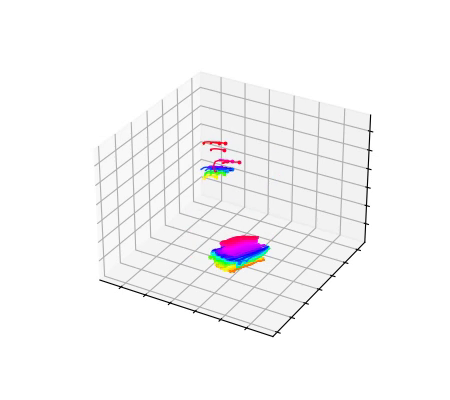} & 
        \includegraphics[clip, height=0.28\linewidth, trim=2.5cm 2cm 2.5cm 2cm]{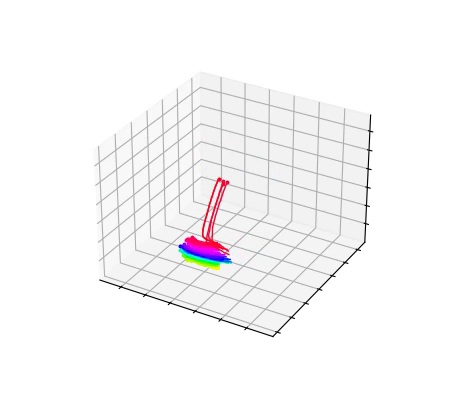} & 
        \includegraphics[clip, height=0.28\linewidth, trim=2.5cm 2cm 2.5cm 2cm]{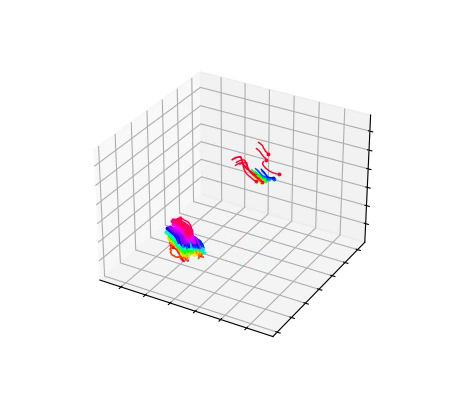} \\ \\

        \includegraphics[width=0.32\linewidth]{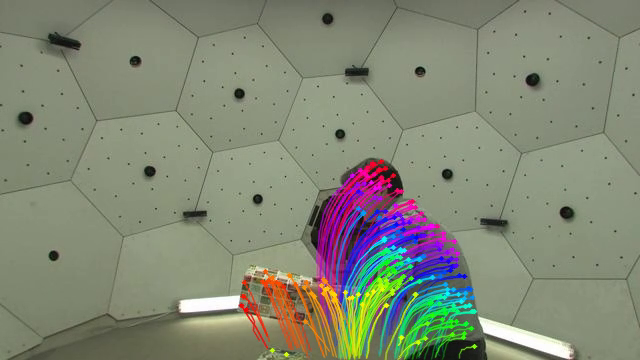} & 
        \includegraphics[width=0.32\linewidth]{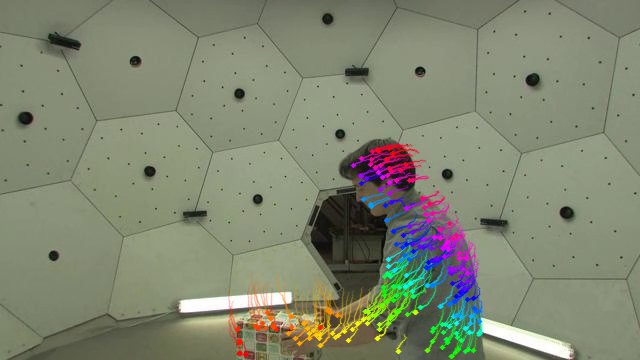} & 
        \includegraphics[width=0.32\linewidth]{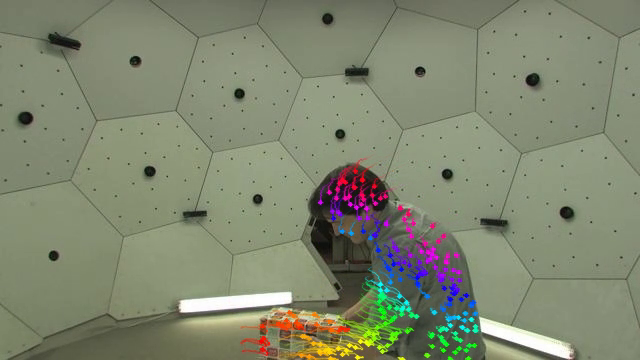} \\
        
        \includegraphics[clip, height=0.28\linewidth, trim=2.5cm 2cm 2.5cm 2cm]{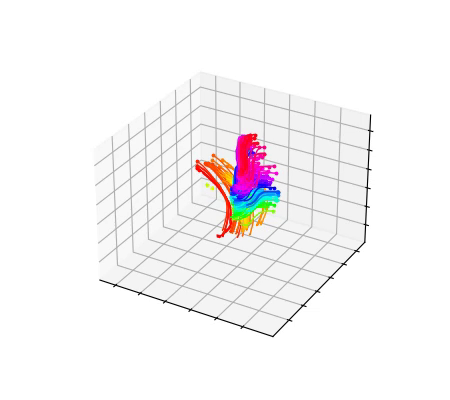} & 
        \includegraphics[clip, height=0.28\linewidth, trim=2.5cm 2cm 2.5cm 2cm]{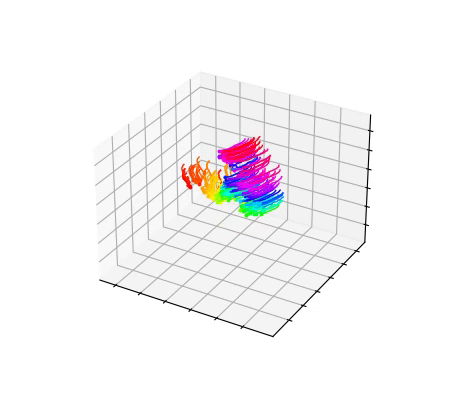} & 
        \includegraphics[clip, height=0.28\linewidth, trim=2.5cm 2cm 2.5cm 2cm]{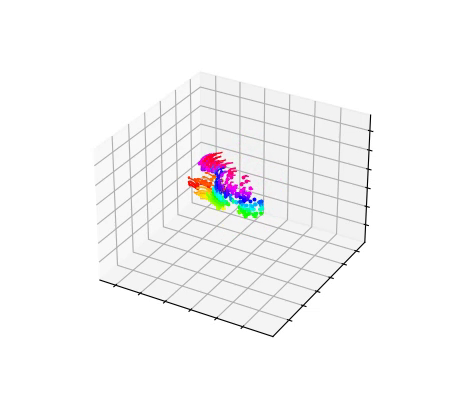} \\
    \end{tabular}
    \caption{Random samples from Panoptic Studio subset in TAPVid-3D: on the top row, we visualize the point trajectories projected into the 2D video frame; on the bottom row, we visualize the metric 3D point trajectories. For each video, we show 3 frames sampled at time step 30, 60 and 90.}
    \label{fig:samples7}
\end{figure}

\begin{figure}[htbp]
    \centering
    \resizebox{\textwidth}{!}{%
    \setlength{\tabcolsep}{2pt}
    \renewcommand{\arraystretch}{1}
    \begin{tabular}{ccc}
    
        \includegraphics[width=0.32\linewidth]{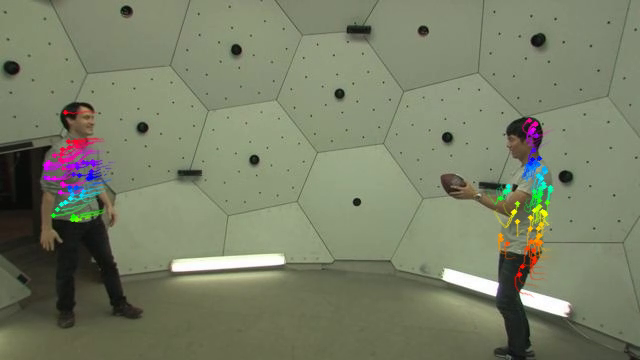} & 
        \includegraphics[width=0.32\linewidth]{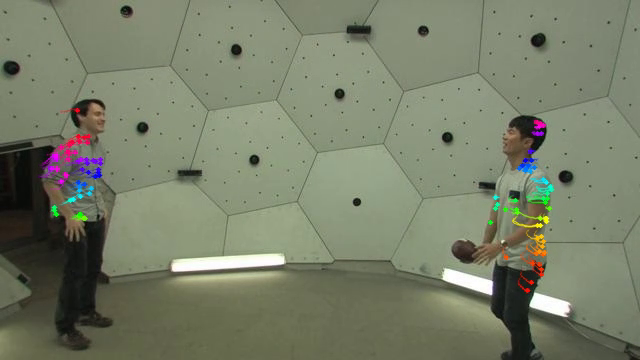} & 
        \includegraphics[width=0.32\linewidth]{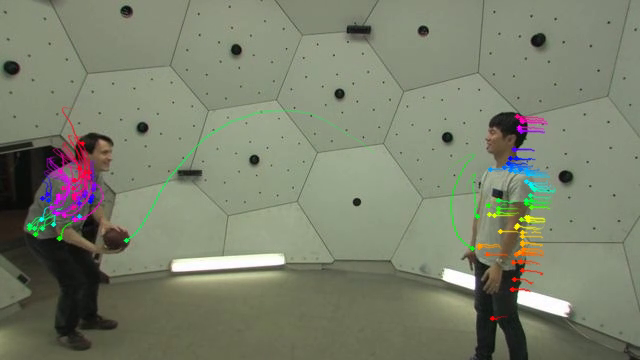} \\
        
        \includegraphics[clip, height=0.28\linewidth, trim=2.5cm 2cm 2.5cm 2cm]{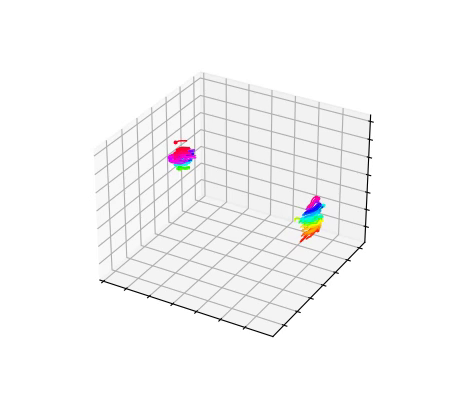} & 
        \includegraphics[clip, height=0.28\linewidth, trim=2.5cm 2cm 2.5cm 2cm]{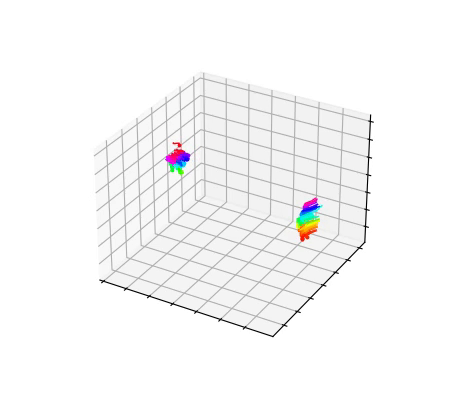} & 
        \includegraphics[clip, height=0.28\linewidth, trim=2.5cm 2cm 2.5cm 2cm]{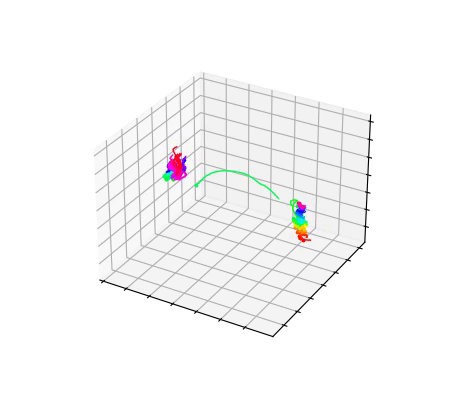} \\ \\
        
        \includegraphics[width=0.32\linewidth]{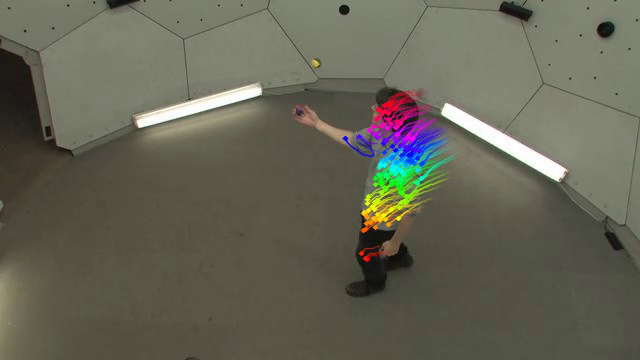} & 
        \includegraphics[width=0.32\linewidth]{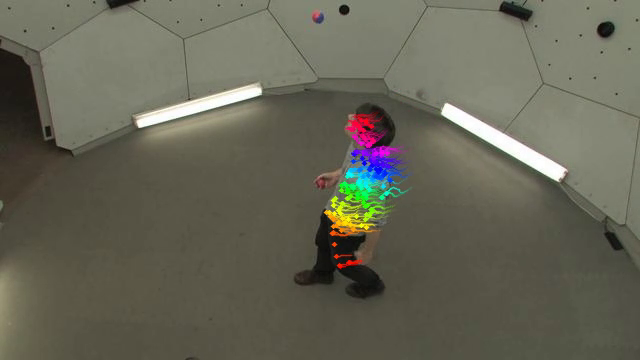} & 
        \includegraphics[width=0.32\linewidth]{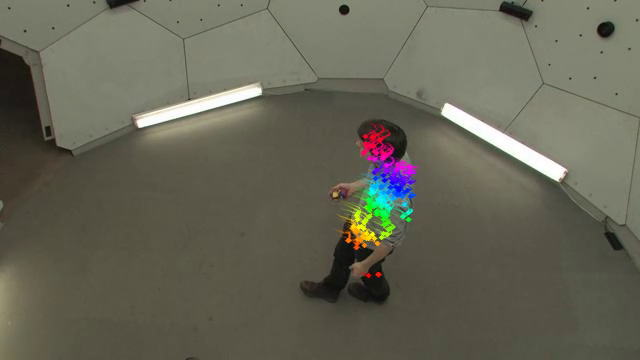} \\
        
        \includegraphics[clip, height=0.28\linewidth, trim=2.5cm 2cm 2.5cm 2cm]{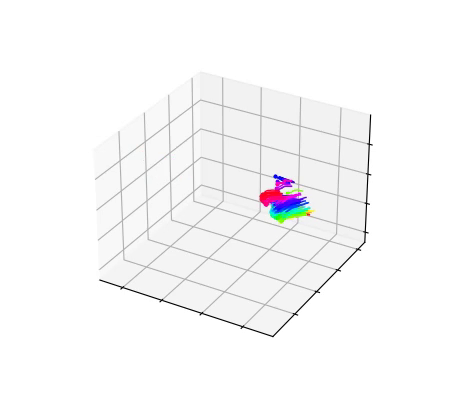} & 
        \includegraphics[clip, height=0.28\linewidth, trim=2.5cm 2cm 2.5cm 2cm]{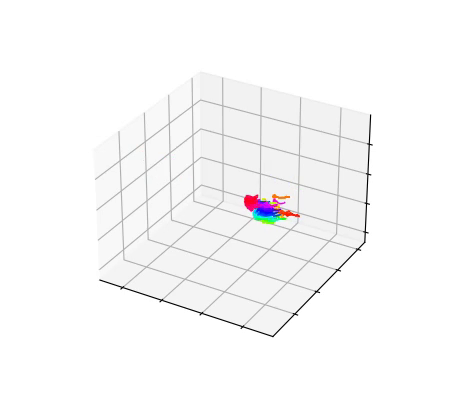} & 
        \includegraphics[clip, height=0.28\linewidth, trim=2.5cm 2cm 2.5cm 2cm]{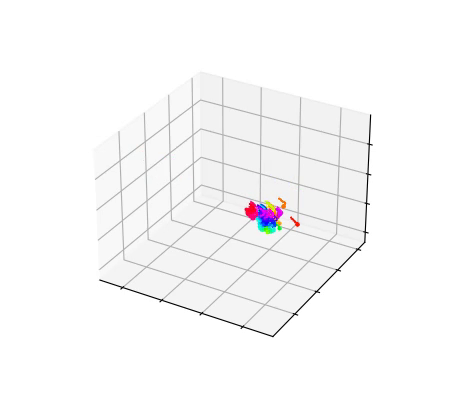} \\
    \end{tabular}
    }
    \caption{Random samples from Panoptic Studio subset in TAPVid-3D (cont'd.): on the top row, we visualize the point trajectories projected into the 2D video frame; on the bottom row, we visualize the metric 3D point trajectories. For each video, we show 3 frames sampled at time step 30, 60 and 90.}
    \label{fig:samples8}
\end{figure}

\begin{figure}[htbp]
    \centering
    \resizebox{\textwidth}{!}{%
    \setlength{\tabcolsep}{2pt}
    \renewcommand{\arraystretch}{1}
    \begin{tabular}{ccc}
        \includegraphics[width=0.32\linewidth]{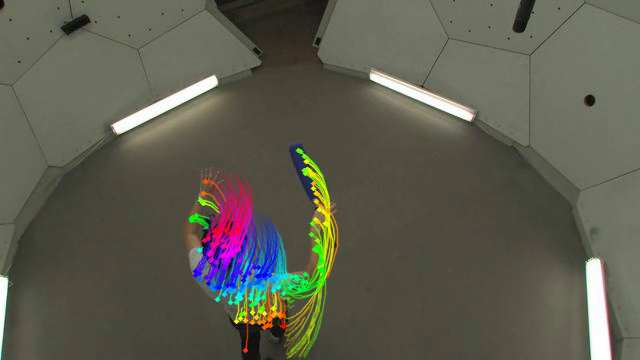} & 
        \includegraphics[width=0.32\linewidth]{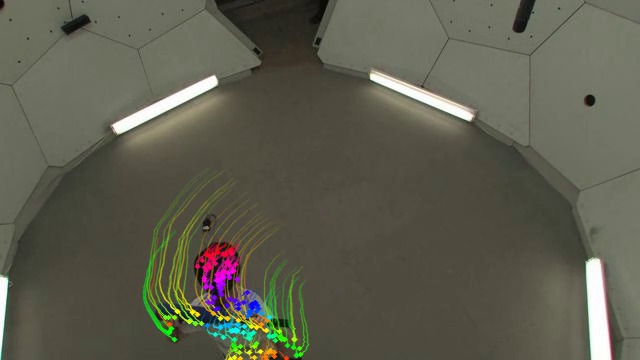} & 
        \includegraphics[width=0.32\linewidth]{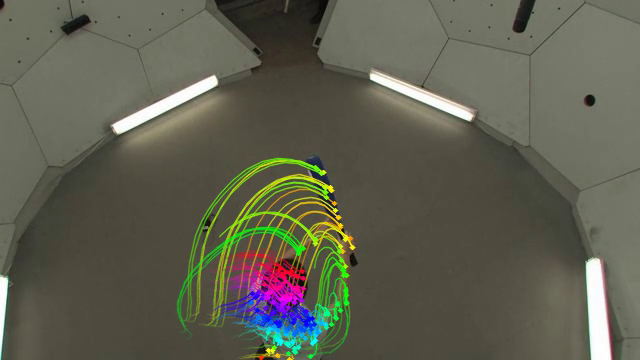} \\
    
        \includegraphics[clip, height=0.28\linewidth, trim=2.5cm 2cm 2.5cm 2cm]{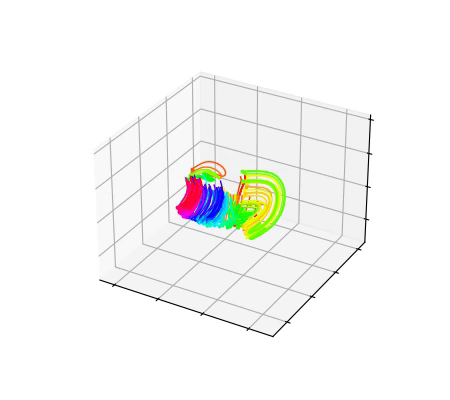} & 
        \includegraphics[clip, height=0.28\linewidth, trim=2.5cm 2cm 2.5cm 2cm]{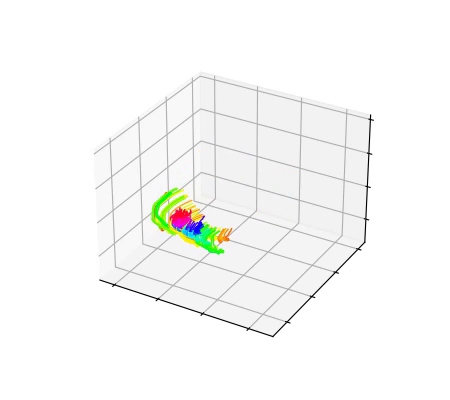} & 
        \includegraphics[clip, height=0.28\linewidth, trim=2.5cm 2cm 2.5cm 2cm]{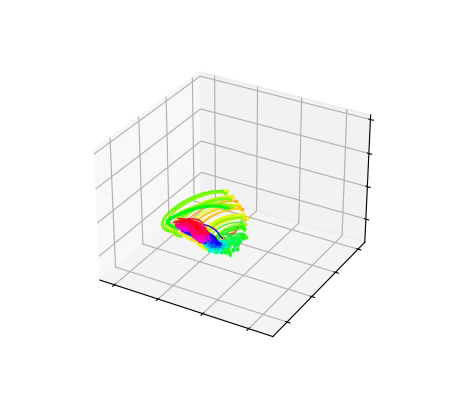} \\ \\
        
        \includegraphics[width=0.32\linewidth]{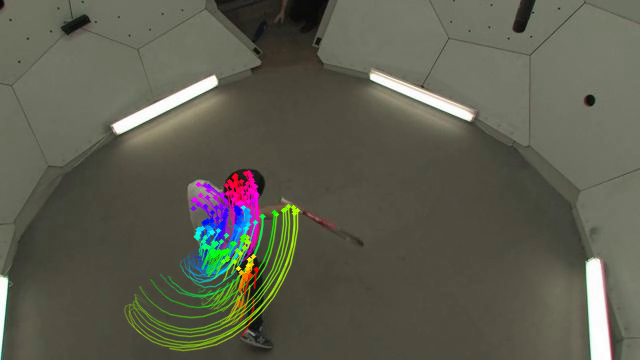} & 
        \includegraphics[width=0.32\linewidth]{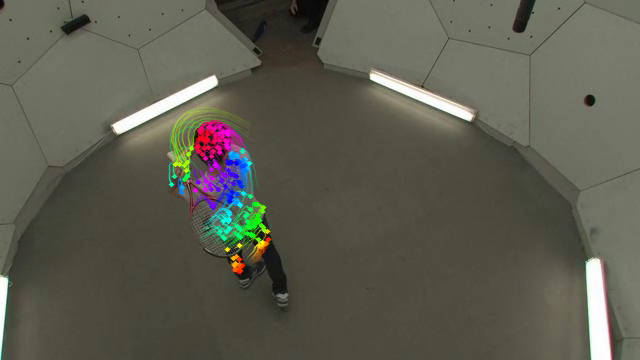} & 
        \includegraphics[width=0.32\linewidth]{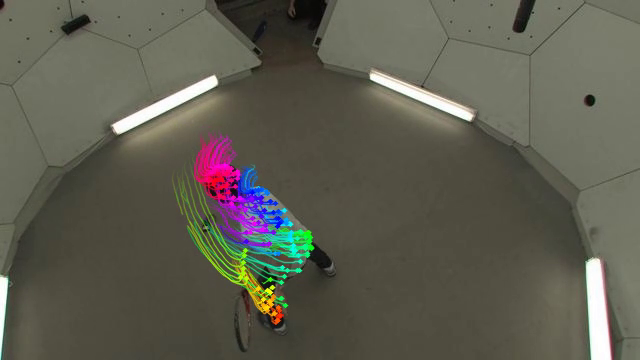} \\
        
        \includegraphics[clip, height=0.28\linewidth, trim=2.5cm 2cm 2.5cm 2cm]{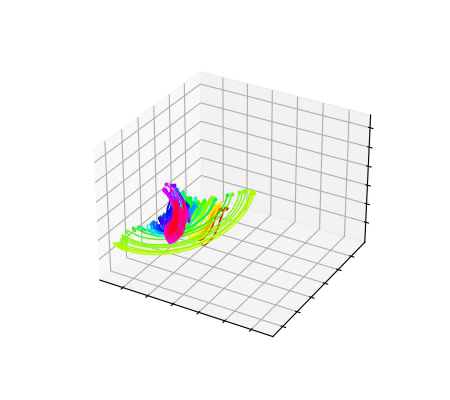} & 
        \includegraphics[clip, height=0.28\linewidth, trim=2.5cm 2cm 2.5cm 2cm]{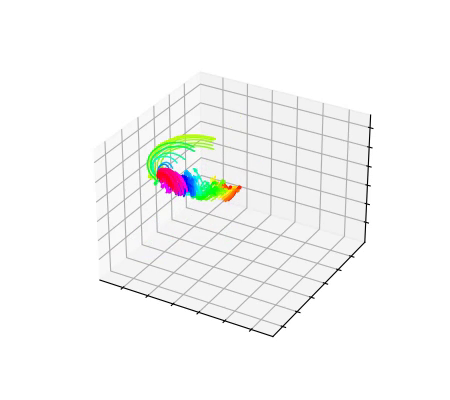} & 
        \includegraphics[clip, height=0.28\linewidth, trim=2.5cm 2cm 2.5cm 2cm]{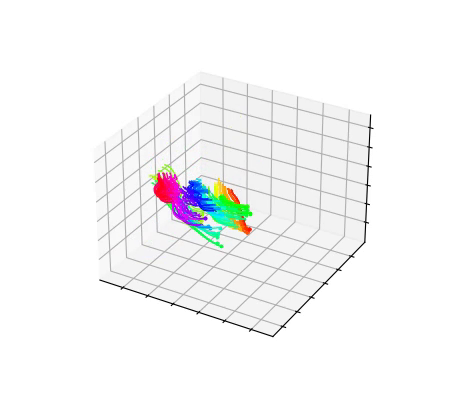} \\
    \end{tabular}
    }
    \caption{Random samples from Panoptic Studio subset in TAPVid-3D (cont'd.): on the top row, we visualize the point trajectories projected into the 2D video frame; on the bottom row, we visualize the metric 3D point trajectories. For each video, we show 3 frames sampled at time step 30, 60 and 90.}
    \label{fig:samples9}
\end{figure}

\end{document}